\documentclass{article}

\usepackage[preprint]{neurips_2023}

\usepackage[utf8]{inputenc} % allow utf-8 input
\usepackage[T1]{fontenc}    % use 8-bit T1 fonts
\usepackage{hyperref}       % hyperlinks
\usepackage{url}            % simple URL typesetting
\usepackage{booktabs}       % professional-quality tables
\usepackage{amsfonts}       % blackboard math symbols
\usepackage{nicefrac}       % compact symbols for 1/2, etc.
\usepackage{microtype}      % microtypography

\usepackage{microtype}
\usepackage{graphicx}
\usepackage{subfigure}
\usepackage{booktabs} % for professional tables
\usepackage[x11names]{xcolor}
\usepackage{tcolorbox}
\usepackage{makecell}
\usepackage{xcolor}

\usepackage{amsmath}
\usepackage{amssymb}
\usepackage{mathtools}
\usepackage{amsthm}

\theoremstyle{plain}

\theoremstyle{definition}

\theoremstyle{remark}

\usepackage[textsize=tiny]{todonotes}

\usepackage{algorithm}
\usepackage{algorithmic}
\usepackage{adjustbox}

\title{Faster Training of Diffusion Models and Improved Density Estimation via Parallel Score Matching}

\author{%
  Etrit Haxholli\\
  Inria\\
  etrit.haxholli@inria.fr
 \And
  Marco Lorenzi\\
  Inria \\
  marco.lorenzi@inria.fr
}

\begin{document}

\maketitle

\begin{abstract}
In Diffusion Probabilistic Models (DPMs), the task of modeling the score evolution via a single time-dependent neural network necessitates extended training periods and may potentially impede modeling flexibility and capacity. To counteract these challenges, we propose leveraging the independence of learning tasks at different time points inherent to DPMs. More specifically, we partition the learning task by utilizing independent networks, each dedicated to learning the evolution of scores within a specific time sub-interval. Further, inspired by residual flows, we extend this strategy to its logical conclusion by employing separate networks to independently model the score at each individual time point. As empirically demonstrated on synthetic and image datasets, our approach not only significantly accelerates the training process by introducing an additional layer of parallelization atop data parallelization, but it also enhances density estimation performance when compared to the conventional training methodology for DPMs.
\end{abstract}

\section{Introduction}

Forward Diffusion Processes (FDPs) represent a category of Markov chains that methodically metamorphose the data distribution into a standard multivariate normal one by incrementally corrupting the data samples. Within the scope of Diffusion Probabilistic Models (DPMs), the machine learning task encompasses the training of a neural network with the objective of emulating the reverse dynamics of this process \cite{DBLP:journals/corr/Sohl-DicksteinW15}.
To this end, different approaches have been developed, notably the denoising \citep{6795935, DBLP:journals/corr/abs-2006-11239} and the sliced score matching loss \citep{JMLR:v6:hyvarinen05a, DBLP:journals/corr/abs-1905-07088}. In both cases, the observed samples are used to train a function approximator in order to model the score, that is, the gradient field of the log-likelihood of the probability density function (pdf) from which the samples originate. 

\begin{figure}[ht]
\centering
\includegraphics[scale=0.12]{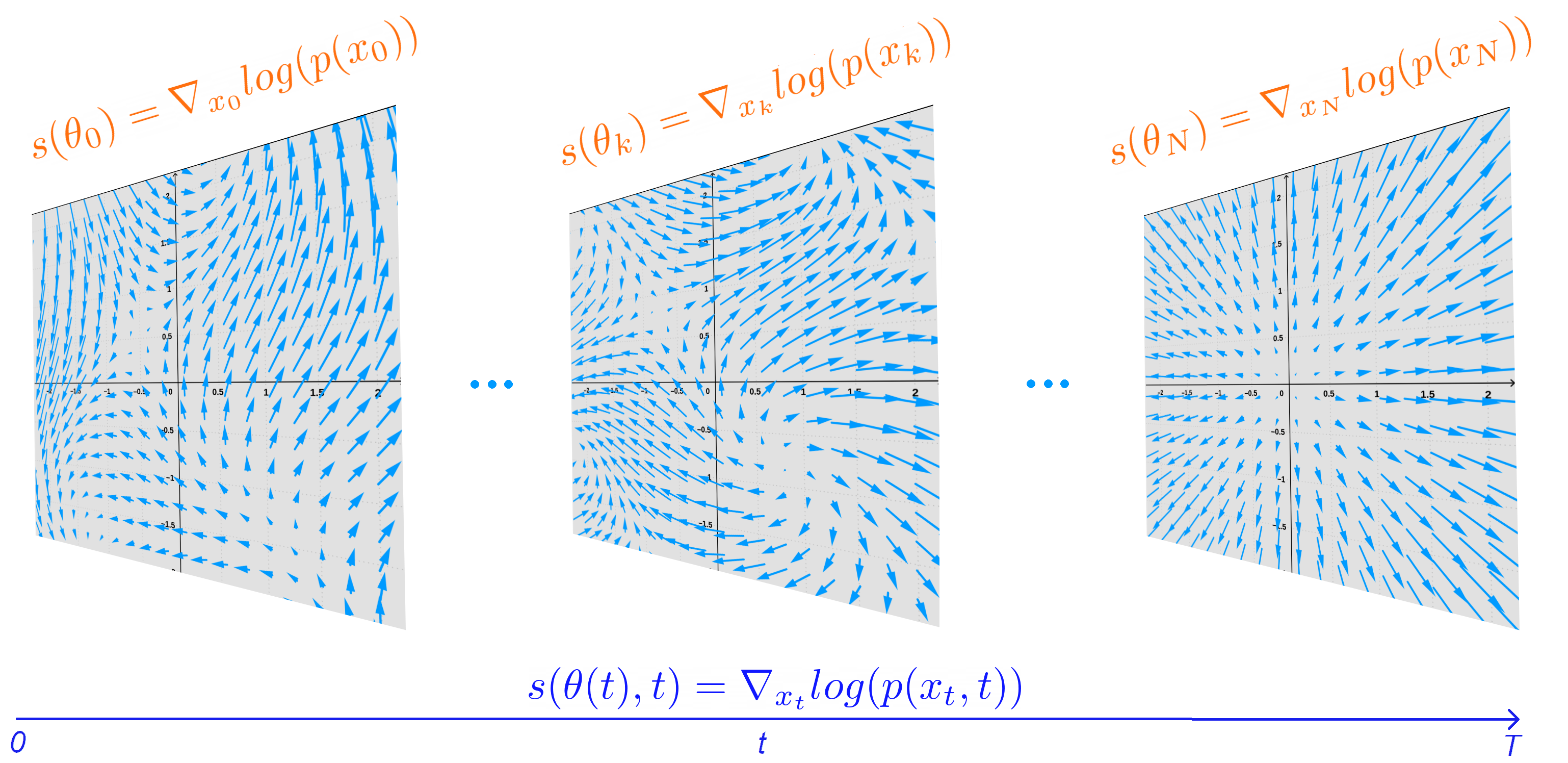}
\caption{Instead of using a time dependent network to learn the scores for all times (below: blue), at each time point we can use a model per score (above: orange), naturally allowing parallel training of diffusion models.}
\label{framework}
\end{figure}

The performance of such models improves as the number of steps in the FDP increases. The optimum is naturally reached at the limit when the number of steps tends to infinity. In this case, the FDP can be depicted as a Stochastic Differential Equation (SDE). This SDE defines a distinct distribution at every temporal point of the diffusion process, with the initial and terminal ones being the data and the standard Gaussian distributions respectively. Given samples from the data distribution, it is possible to efficiently generate samples from the distribution at any specified time $t_i$. These generated samples can then be utilized to train a neural network to approximate the score at time $t_i$. The common approach \cite{DBLP:journals/corr/abs-2006-11239, song2021scorebased, DBLP:journals/corr/abs-2112-10752} is to train a single time-varying neural network to model all scores, i.e., to model the evolution of the score (Figure \ref{framework}).

The Fokker-Planck equation establishes the connection between SDE diffusion models and continuous normalizing flows \citep{song2021scorebased}. Indeed, the estimated scores provide the dynamics of the continuous normalizing flow (CNF) which describes the evolution of the distribution determined by the SDE. Leveraging on this connection, the CNF framework can be deployed to generate data, or to estimate the likelihood of unobserved points through the application of the instantaneous change of variable.

Despite the inherent equivalence and functionality between SDE diffusion processes and CNFs, their optimization techniques exhibit differences. CNFs are trained by maximizing likelihood, a methodology that is suboptimal for numerous reasons. A key reason is that normalizing flows, being designed as a sequence of transformations, necessitate the retention of the entire sequence in memory during training, as it is infeasible to optimize a step in the sequence independently from the rest. While CNFs alleviate the memory bottleneck via the adjoint method, the continuity requirement dictates that all vector fields at all time points must be learned by a singular time-dependent neural network, which impinges upon the transformation's flexibility. The composition of multiple CNFs together is rendered impracticable due to memory and computational constraints, as the sequential nature of the composition, as before, implies that each additional CNF introduces a new set of parameters and extends the training duration. In this context, each CNF is referred to as a CNF block.

The aforementioned points underscore the impracticability of independent learning of vector fields in the CNF approach. To expedite training and augment model flexibility, we leverage the intrinsic property of DPMs that allows scores at different time points to be optimized independently. Consequently, instead of training a single time-varying U-Net to model the scores of the distributions at all time points, we partition the training task by dividing the integration time interval of the SDE into smaller sub-intervals. For each sub-interval, we employ a time-dependent U-Net to model the evolution of scores for the distributions defined within that sub-interval. This strategy is referred to as time-varying parallel score matching (TPSM). Each such U-Net corresponds to a CNF block within the CNF framework, and their composition provides the complete evolution of distribution as determined by the FDP.

With an increasing number of such sub-intervals, the task assigned to each U-Net is simplified, and in the limit, each time sub-interval converges to a point. Consequently, in addition to the previously described strategy, we train a multitude of smaller networks, which are not time-dependent, such that each network learns the score at a single time point of the diffusion process (Figure \ref{framework}). This methodology is designated as Discrete Parallel Score Matching (DPSM).

We evaluate our Parallel Score Matching (PSM) methodologies on 2D data, in addition to image datasets such as CIFAR-10, CelebA 64$\times$64, and ImageNet 64$\times$64, \cite{liu2015faceattributes, deng2009imagenet}. The findings attest that our approach not only enhances the log-likelihood results, but also expedites the training process, without incurring penalties related to memory or inference time.

\section{Background and Related Work}

\subsection{Normalizing Flows (NFs)}
\textbf{Normalizing Flows:} 
A Normalizing Flow \citep{cms/1266935020, https://doi.org/10.1002/cpa.21423, https://doi.org/10.48550/arxiv.1505.05770, https://doi.org/10.48550/arxiv.1410.8516} is a transformation defined as a sequence of diffeomorphisms that converts a base probability distribution (e.g., a standard normal) into another distribution by warping the domain on which they are defined. Let $\boldsymbol{Z}$ be a random variable, and define $\boldsymbol{X}=g(\boldsymbol{Z})$, where $g$ is a diffeomorphism with inverse $h$. If we denote their probability density functions by $f_{\boldsymbol{Z}}$ and $f_{\boldsymbol{X}}$, the change of variable theorem states that:
%\begin{equation}\label{s1}
%f_X(x)=f_\boldsymbol{Z}(\boldsymbol{z})|\frac{d\boldsymbol{z}}{dx}|=f_\boldsymbol{Z}(\boldsymbol{z})|\frac{1}{\frac{dx}{d\boldsymbol{z}}}|=f_\boldsymbol{Z}(\boldsymbol{z})|\frac{dg(\boldsymbol{z})}{d\boldsymbol{z}}|^{-1}=f_X(g(\boldsymbol{z}))
%\end{equation}
%and
\begin{equation}\label{s2}
f_{\boldsymbol{X}}(\boldsymbol{x})=f_{\boldsymbol{Z}}(\boldsymbol{z})|\frac{d\boldsymbol{z}}{d\boldsymbol{x}}|=f_{\boldsymbol{Z}}(h(\boldsymbol{x}))|\frac{d h(\boldsymbol{x})}{d\boldsymbol{x}}|.
\end{equation}
The objective is the optimization of parameters of $h$ to maximize the likelihood of sampled data points $\boldsymbol{x}$.
After training, one can input any test point ${\boldsymbol{x}}$ on the RHS of Equation \ref{s2}, and calculate its likelihood. For the generative task, being able to easily recover $g$ from $h$ is essential, as the generated point ${\boldsymbol{x}}_g$ will take form ${\boldsymbol{x}}_g$=$g(\boldsymbol{z}_s)$, where $\boldsymbol{z}_s$ is a sampled point from the base distribution $f_{\boldsymbol{Z}}$.
\newline
For increased modeling flexibility, we can use a chain (flow) of transformations, $\boldsymbol{z}_i=g_i(\boldsymbol{z}_{i-1}), i\in[n]$. The interested reader can find a more in-depth review of normalizing flows in \citep{Kobyzev_2021} and \citep{JMLR:v22:19-1028}.

\textbf{Neural ODE Flows:}  Neural ODEs \citep{https://doi.org/10.48550/arxiv.1806.07366} are continuous generalizations of residual networks:
\begin{equation*}
\boldsymbol{x}_{t_{i+1}}=\boldsymbol{x}_{t_i}+\epsilon f(\boldsymbol{x}_{t_i}, t_i, \boldsymbol{\theta})
\end{equation*}
\begin{equation}\label{neural_ODEs}
\rightarrow \boldsymbol{x}(t)=\boldsymbol{x}(0) +\int_0^t f(\boldsymbol{x}(\tau),\tau,\boldsymbol{\theta}) d\tau \text{, as } \epsilon \rightarrow 0,
\end{equation}

where for a network with finite weights, injectivity is guaranteed by the Picard–Lindelöf theorem. In \citep{ https://doi.org/10.48550/arxiv.1806.07366}, the expression for the instantaneous change of variable is derived, which enables one to train continuous normalizing flows and perform likelihood estimation:
\begin{equation}\label{instantaneous_change}
    \log{p(\boldsymbol{x}(0))}=\log{p(\boldsymbol{x}(T))}+\int_0^T tr{\frac{\partial  f(\boldsymbol{x}(t), t, \boldsymbol{\theta})}{\partial \boldsymbol{x}(t)}dt},
\end{equation}
where $\boldsymbol{x}(0)$ represents a sample from the data. Such models are better known as Continuous Normalizing Flows (CNFs).

\textbf{Invertible Residual Flows (IRFs):}  These models \citep{pmlr-v97-behrmann19a} are similar to the discretized version of Continuous Normalizing Flows:
\begin{equation*}
\boldsymbol{x}_{t_{i+1}}=\boldsymbol{x}_{t_i}+f(\boldsymbol{x}_{t_i}, \boldsymbol{\theta_{t_i}})
\end{equation*}

It can be noticed that in this case the output of the residual block is not scaled before being added to the input, and furthermore, each block is comprised of a different set of trainable parameters, instead of allowing the behaviour of the model to evolve via a time input. As injectivity cannot be guaranteed via the Picard–Lindelöf theorem, bijectivity has to be enforced by imposing the contraction condition $Lip(f(\boldsymbol{x}_{t_i}, \boldsymbol{\theta_{t_i}}))<1$, where $Lip(f(\boldsymbol{x}_{t_i}, \boldsymbol{\theta_{t_i}}))$ is the Lipschitz constant of $f(\boldsymbol{x}_{t_i}, \boldsymbol{\theta_{t_i}})$.

\subsection{Diffusion Probabilistic Models (DPMs)}

\textbf{Discrete Diffusion Models:}
A forward diffusion process or diffusion process \citep{DBLP:journals/corr/Sohl-DicksteinW15} is a fixed Markov chain that
gradually adds Gaussian noise to the data according to a schedule $\{b_t|t \in [n] \}$:
\begin{equation}\label{forward_diff}
q(\boldsymbol{x}_{t}|\boldsymbol{x}_{t-1}):= \mathcal{N}(\boldsymbol{x}_{t-1}\sqrt{1-b_{t}}, b_{t}\boldsymbol{I}).
\end{equation}
Such a process transforms the data distribution $p(\boldsymbol{x}_0)$ into a standard multivariate normal distribution. The goal is to approximate the reverse process via
\begin{equation}\label{reverse_diff}
p(\boldsymbol{x}_{t-1}|\boldsymbol{x}_{t}):= \mathcal{N}(\mu_{\boldsymbol{\theta}}( \boldsymbol{x}_{t},t), \boldsymbol{\Sigma}(t)),
\end{equation}
 that converts the standard multivariate normal distribution into the data distribution. To this end, for each step $t$, the KL divergence between $q(\boldsymbol{x}_{t-1}|\boldsymbol{x}_{t}, \boldsymbol{x}_{0})$ and $p(\boldsymbol{x}_{t-1}|\boldsymbol{x}_{t})$ is minimized, which amounts to minimizing
\begin{equation}
\mathbb{E}_{\boldsymbol{x_{0}},\boldsymbol{x_{t}}}||\mu_{\boldsymbol{\theta}}( \boldsymbol{x}_{t},t)-\mu(\boldsymbol{x}_{0}, \boldsymbol{x}_{t})||^2,
\end{equation}
where $\mu_{\boldsymbol{\theta}}( \boldsymbol{x}_{t},t)$ is a neural network, while $\mu(\boldsymbol{x}_{0}, \boldsymbol{x}_{t})$ is the mean of $q(\boldsymbol{x}_{t-1}|\boldsymbol{x}_{t}, \boldsymbol{x}_{0})$.
\subsection{SDE Diffusion Models and their CNF Representation}

For an $\boldsymbol{\varepsilon} \sim \mathcal{N}(\boldsymbol{0},\boldsymbol{I})$, the forward diffusion process defined in Equation \ref{forward_diff} can be written as 
\begin{equation}\label{forward_diff2}
\boldsymbol{x}_{t}=\boldsymbol{x}_{t-1}\sqrt{1-b_{t}}+ \sqrt{b_{t}} \boldsymbol{\varepsilon}.
\end{equation}
As derived in \citep{song2021scorebased}, the continuous counterpart of this process takes the form
\begin{equation}\label{cont_forward_diff}
d\boldsymbol{x}(t)=-\frac{1}{2}b(t)\boldsymbol{x}(t)dt+ \sqrt{b(t)} d\boldsymbol{w},
\end{equation}
\begin{flalign}\label{cont_conditional_forward_prob}
&\text{thus }q(\boldsymbol{x}_{t}|\boldsymbol{x}_{0})= \mathcal{N}(\boldsymbol{x}_{0}\mu_t, \sigma_t^2 \boldsymbol{I}) \text{, where } \mu_t=e^{-\frac{1}{2}\int_0^t b(s)ds} \text{ and } \sigma_t=\sqrt{(1-e^{-\int_0^t b(s)ds})}.&
\end{flalign}
The evolution of the probability density function of the data, as dictated by the FDP, is identical to the evolution dictated by the following ODE transformation:
\begin{equation}\label{smooth_forward_diff}
d\boldsymbol{x}(t)=-\frac{1}{2}b(t)[\boldsymbol{x}(t)+\nabla_{\boldsymbol{x}(t)}\log{p_t(\boldsymbol{x}(t))}]dt=f_t(\boldsymbol{x}(t))dt.
\end{equation}
Knowing $f_t$ allows us to perform data generation and likelihood estimation, through the CNF framework. It can be observed that the only unknown in $f_t$, is the score $\nabla_{\boldsymbol{x}(t)}\log{p_t(\boldsymbol{x}(t))}$. This quantity can be modelled using a neural network $\boldsymbol{s}_{\theta}(\boldsymbol{x}(t),t)$ trained through the MSE denoising loss :
\begin{equation}
\mathbb{E}_{\boldsymbol{x}_{0},\boldsymbol{\varepsilon}}||\boldsymbol{s}_{\theta}(\boldsymbol{x}_{0}\mu_t+\sigma_t \boldsymbol{\varepsilon},t)-(-\frac{\boldsymbol{\varepsilon}}{\sigma_t})||^2,
\end{equation}  

derived in \cite{DBLP:journals/corr/abs-2006-11239, DBLP:journals/corr/abs-2107-00630}. A more detailed explanation of discrete diffusion models and their continuous counterpart can be found in Appendix \ref{AppendixE}.
\section{PSM Framework and Relation to Piece-wise Continuous Flows and IRFs}

In the context of continuous normalizing flows, given the forward integration, the expression $\boldsymbol{x}(t)=\boldsymbol{x}(0)+\int_0^T f_{t} dt$ is comprised of all vector fields $f_t=f(\boldsymbol{x}(t),\boldsymbol{\theta}, t)$, parameterized by the network's trainable parameters, as well as the time input. Increasing model capacity by using a distinct neural network to learn a vector field at each time $t$ is not feasible, as the entire function $\boldsymbol{x}(t_N)=\boldsymbol{x}(\boldsymbol{x}_0, f_{t_1}(\boldsymbol{\theta}_{t_1}),f_{t_2}(\boldsymbol{\theta}_{t_2}),...,f_{t_N}(\boldsymbol{\theta}_{t_T}))$ would need to be maintained in memory, and the transformation may not necessarily be continuous. Memory constraints also surface when employing multiple CNF blocks within a piece-wise continuous flow, as each block $\int_{t_{i-1}}^{t_i} f(\boldsymbol{x}(\tau),\boldsymbol{\theta}_i, \tau) d\tau$ introduces a new set of parameters $\boldsymbol{\theta}_i$. Given that these transformations are interconnected in a  chain, they must coexist in memory during the maximum likelihood optimization procedure. Moreover, continuous normalizing flows present a formidable optimization task, as the time-dependent network $f(\boldsymbol{x}(t),\boldsymbol{\theta}, t)$ is obligated to generate an evolution process for the data distribution that culminates in a standard normal distribution. The loss function further adds to the complexity, as not only it contains the end point of the integration path $\boldsymbol{x}(T)=\boldsymbol{x}(0)+\int_0^T f(\boldsymbol{x}(t),\boldsymbol{\theta}, t) dt$, but also the integral of the divergence along that integration path $\int_0^T tr{\frac{\partial  f(\boldsymbol{x}(t), t, \boldsymbol{\theta})}{\partial \boldsymbol{x}(t)}dt}$.
\newline
Contrarily, diffusion models parameterized by time-dependent neural networks ameliorate the complexity of learning the evolution process. This is because the forward diffusion delineates the time-based evolution of the data distribution, simplifying the framework's task to merely modeling the already defined vector fields. However, if parameterized by a single time-varying network, these models still grapple with limited flexibility due to a single set of parameters (model) being tasked with modeling all the vector fields defined by the FDP. Moreover, the optimization of a vector field at time $t$ is dependent on the optimization of other vector fields at various time points, which hampers the speed of the training process. This standard approach (SA) in Diffusion Probabilistic Models (DPMs) is outlined in Algorithm \ref{TDUM} in Appendix \ref{AppendixE} for comparison, and is abbreviated as SA-DPM.

\begin{tcolorbox}[width=1\linewidth,colback={Snow1},title={Disadvantages of SA in DPMs},colbacktitle=Snow2,coltitle=black]
    1) The flow's adaptability over time is limited due to the employment of a single time-varying network.
    \\
    2) The optimization of score approximation at any given time $t_i$ remains influenced by the optimization of score approximation at any other time $t_j$.
\end{tcolorbox}

\begin{table}[t]
\caption{A comparison of the properties of Parallel Score Matching (PSM) approaches  with other generative diffeomorphism-based frameworks.}
\label{table1}
\vskip 0.15in
\begin{center}
\begin{small}
\begin{sc}
\begin{tabular}{lcccr}
\toprule
Method: & CNF & SA-DPM & PSM-DPM \\
\midrule
\makecell{Computationally  Demanding Loss}     & \textcolor{red}{$\surd$} & \textcolor{green}{$\boldsymbol{\times}$} & \textcolor{green}{$\boldsymbol{\times}$} \\
 & & & \\
\makecell{Memory Bottleneck}     & \textcolor{red}{$\surd$} & \textcolor{green}{$\boldsymbol{\times}$} & \textcolor{green}{$\boldsymbol{\times}$} \\
 & & & \\
\makecell{Limited  Variability In Time}& \textcolor{red}{$\surd$} & \textcolor{red}{$\surd$} & \textcolor{green}{$\boldsymbol{\times}$} \\
& & & \\
\makecell{Necessity To Devise  A diffeomorphism}& \textcolor{red}{$\surd$} & \textcolor{green}{$\boldsymbol{\times}$} & \textcolor{green}{$\boldsymbol{\times}$} \\
& & & \\
\makecell{Score Optimization  Interdependency}& \textcolor{red}{$\surd$} & \textcolor{red}{$\surd$} & \textcolor{green}{$\boldsymbol{\times}$} \\
\bottomrule
\end{tabular}
\end{sc}
\end{small}
\end{center}
\vskip -0.1in
\end{table}  

As shown in Appendix \ref{AppendixD}, the scores in a diffusion process evolve continuously, with the given PDE dynamics:
\begin{equation*}
     \frac{\partial s(\boldsymbol{x},t)}{\partial t} = \frac{1}{2}b(t)\nabla_{\boldsymbol{x}} tr\frac{\partial s(\boldsymbol{x},t)}{\partial \boldsymbol{x}}+\frac{1}{2}b(t)\frac{\partial s(\boldsymbol{x},t)}{\partial \boldsymbol{x}} (\boldsymbol{x}+s(\boldsymbol{x},t)).
\end{equation*}

\begin{figure}[b]
    \centering
    \scriptsize
    \vspace{-5.5mm}
    \begin{minipage}[t]{0.495\textwidth}
        \begin{algorithm}[H]
            \caption{\\Time-varying Parallel Score Matching (TPSM)}
            \label{TDMM}
            \begin{algorithmic}
                \STATE {\bfseries Input:} data $\{\boldsymbol{x}_1,...,\boldsymbol{x}_n\}$, batch-size p
                \STATE Split $[0,1]$ into $\cup_{i=0}^{N-1}[t_i,t_{i+1}]$
                \FOR{$i=0$ {\bfseries to} $N-1$, in parallel}
                \STATE \emph{Train model $\boldsymbol{s}_{\boldsymbol{\theta}_i}$}
                \REPEAT
                \STATE uniformly sample $\tau_1$, ..., $\tau_p$ from $[t_i,t_{i+1}]$
                \STATE sample $\boldsymbol{x}^0_{\pi(1)}$, ..., $\boldsymbol{x}^0_{\pi(p)}$ from $\{\boldsymbol{x}_1,...,\boldsymbol{x}_n\}$
                \STATE generate $\boldsymbol{x}^{\tau_1}_{\pi(1)}$, ..., $\boldsymbol{x}^{\tau_p}_{\pi(p)}$ using Eq. \ref{cont_conditional_forward_prob}
                \STATE minimize $ \Sigma_j ||\boldsymbol{\varepsilon}_{\boldsymbol{\theta}_i}(\boldsymbol{x}^{\tau_j}_{\pi(j)}(\boldsymbol{\varepsilon_j}),\tau_j)-(\boldsymbol{\varepsilon_j})||^2$
                \UNTIL convergence
                \ENDFOR
            \end{algorithmic}
        \end{algorithm}
    \end{minipage}
    \hfill
    \begin{minipage}[t]{0.495\textwidth}
        \begin{algorithm}[H]
            \caption{\\Discrete Parallel Score Matching (DPSM)}
            \label{DDMM}
            \begin{algorithmic}
               \STATE {\bfseries Input:} data $\{\boldsymbol{x}_1,...,\boldsymbol{x}_n\}$, batch-size p
               \STATE Discretize $[0,1]$ into $\{t_0=0, t_1, ...,t_N=1\}$
               \FOR{$i=1$ {\bfseries to} $N$, in parallel}
               \STATE \emph{Train model $\boldsymbol{s}_{\boldsymbol{\theta}_i}$}
               \REPEAT
               \STATE uniformly sample $t_i$ from $\{t_0, ...,t_N\}$
               \STATE sample $\boldsymbol{x}^0_{\pi(1)}$, ..., $\boldsymbol{x}^0_{\pi(p)}$ from $\{\boldsymbol{x}_1,...,\boldsymbol{x}_n\}$
               \STATE generate $\boldsymbol{x}^{t_i}_{\pi(1)}$, ..., $\boldsymbol{x}^{t_i}_{\pi(p)}$ using Eq. \ref{cont_conditional_forward_prob}
               \STATE minimize $ \Sigma_j ||\boldsymbol{\varepsilon}_{\boldsymbol{\theta}_i}(\boldsymbol{x}^{t_i}_{\pi(j)}(\boldsymbol{\varepsilon_j}))-(\boldsymbol{\varepsilon_j})||^2$
               \UNTIL convergence
               \ENDFOR
             \end{algorithmic}
        \end{algorithm}

    \end{minipage}
\end{figure}

 Given that the distribution at each time $t_i$ and its associated score are intrinsically determined by the data distribution and the Forward Diffusion Process (FDP), we can expedite training by concurrently learning the scores $\nabla_{\boldsymbol{x}_{t_i}} \log p(\boldsymbol{x}_{t_i}, t_i)$ through each $s_{\boldsymbol{\theta}_{i}}(\boldsymbol{x}_{t_i})$ at all times $t_i$. If our models proficiently approximate these scores, the transformation resulting from combining these models will be smooth.

This instigates partitioning the diffusion process, specifically, the interval $t\in [0,1]$ is divided into $\bigcup_{i=0}^{N-1}[t_i,t_{i+1}]$. Thereby, it remains feasible to train a time-dependent neural network $s_{\boldsymbol{\theta_i}}(\boldsymbol{x}_{t},t)$ to acquire scores $\nabla_{\boldsymbol{x}_{t}} \log p(\boldsymbol{x}_{t}, t)$ for time $t \in [t_i,t_{i+1}]$. Such an approach diminishes the complexity of tasks for each model, bolstering modeling capability, whilst preserving time continuity. Moreover, given the parallelizability of training, a solitary model is loaded in memory per device during training, and post-training it can be stored on a disk and recalled at testing. The training procedure of this framework is described in Algorithm \ref{TDMM}. In this case, each score modeling network $s_{\boldsymbol{\theta_i}}(\boldsymbol{x}_{t},t)$ corresponds to a CNF block in the continuous normalizing flow representation.

Elevating this approach to its apex, the duration of each interval approaches $0$, thus we utilize a single network to model the score per time-point. This framework corresponds to an invertible residual flow with infinitesimal scaling, as invertibility and differentiability are guaranteed by the fact that the learned transformation approximates the diffusion process. A description of this training procedure is given in Algorithm \ref{DDMM}. 

A comparison of the properties of Parallel Score Matching (PSM) with other generative diffeomorphism-based frameworks can be found in Table  \ref{table1}. Details about the generation procedure and likelihood estimation in the PSM framework can be found in Appendix \ref{AppendixA}.

\section{Experiments}

We contrast the density estimation performance of DPMs trained via parallel score matching and those trained through the standard approach (SA-DPM). We conduct experiments on 2D data of toy distributions, alongside standard benchmark datasets including CIFAR-10, CelebA, and ImageNet.

For the 2D toy distribution data scenario, we solely compare the TPSM method against the baseline, namely, SA-DPM. The batch size across all experiments is $512$, and we employ a learning rate of $10^{-3}$. Each network was trained on a single CPU of equivalent performance and shares the same architecture in both approaches. This network is an MLP comprised of three hidden layers, wherein the non-linearity is provided by the ELU activation function.

Concerning standard image benchmark datasets, within the TPSM approach, we adopt the torch implementation \cite{ddpmtorch} of the network originally proposed in \citep{DBLP:journals/corr/abs-2006-11239} which, for the purpose of equitable comparison, is also utilized in SA-DPM. Similarly, within the DPSM approach, we opt for the even simpler basic U-Net, which originally presented the U-Net architecture in \citep{DBLP:journals/corr/RonnebergerFB15}. For all models, we implement a batch size of $32$ on ImageNet and CelebA, and a batch size of $128$ for CIFAR-10. In DPSM we adopt a learning rate of $10^{-3}$, while in TPSM and SA-DPM this is reduced to $2\times 10^{-4}$. Each neural network across all three methods was trained on a single GPU with RTX-2080 level of performance. As in the 2D data experiments, we utilize the ELU activation function.

At no stage do we employ data parallelization, which constitutes an auxiliary parallelization strategy that complements the methodology presented in this paper.

\subsection{Toy 2D Datasets}\label{toydata}

\begin{figure*}[t]
\center
\includegraphics[width=1\textwidth]{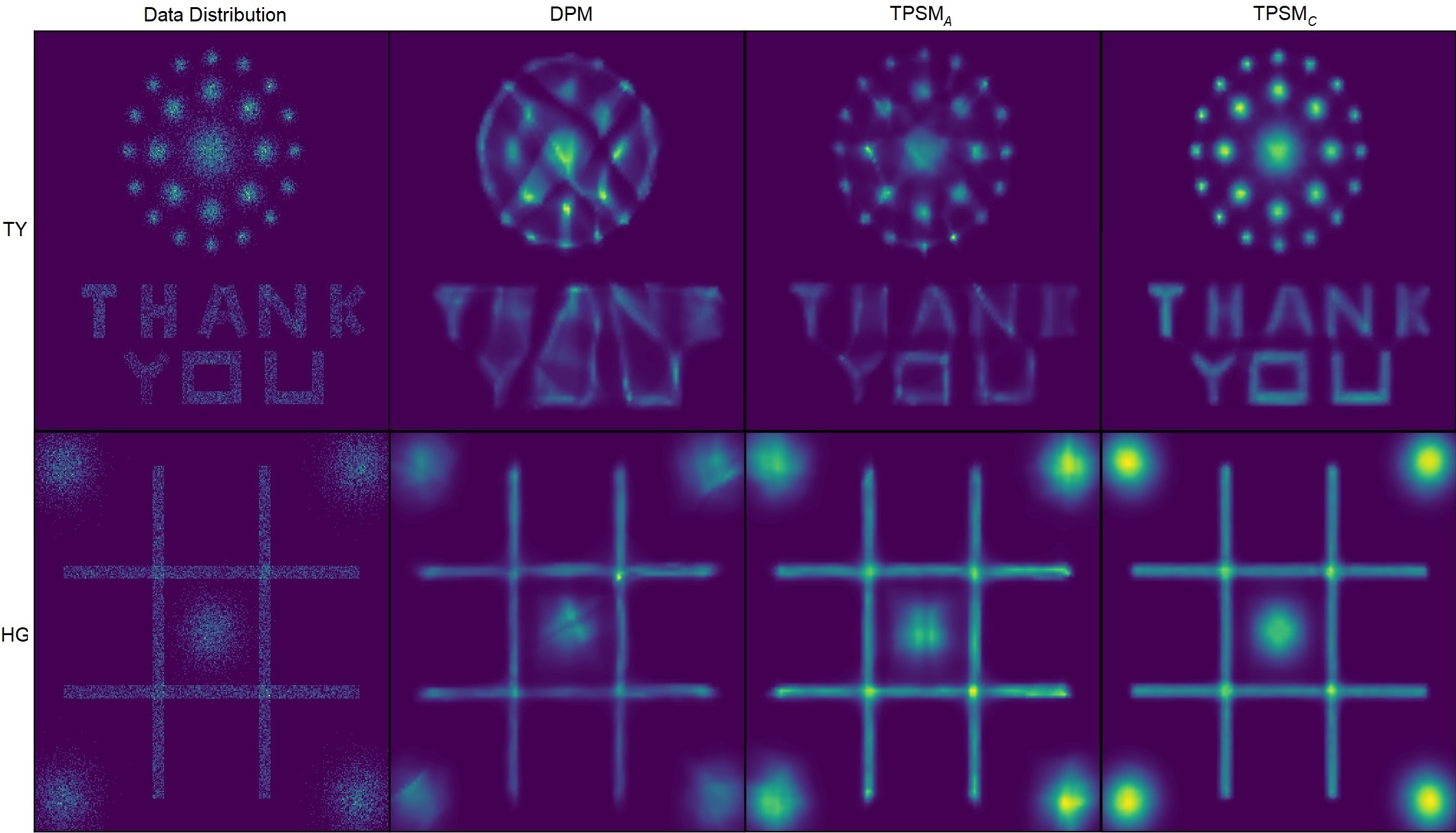}
\caption{Probability density modeling capabilities of SA (second column), TPSM$_A$ (third column) and TPSM$_C$ (fourth column) on 2D data of toy distributions. The performance of TPSM$_B$ is shown in Appendix \ref{AppendixG}.}
\label{toy2dimage}
\end{figure*}

To visually discern the distribution modeling capabilities between the baseline and the TPSM approach, we initially test both frameworks on 2D toy data. The two distributions we examine are TY (refer to the upper row of Figure \ref{toy2dimage}) and HG (lower row of Figure \ref{toy2dimage}). The time-varying network deployed is identical in both methods and for both distributions. More specifically, it is an MLP with the subsequent structure: $4\rightarrow100\rightarrow150\rightarrow100\rightarrow2$. It merits mention that this model's input is 4-dimensional, as the model's time variability is enabled by appending the time component $t$ to the 2D data input $\boldsymbol{x}(t)$. The output is 2-dimensional, as it predicts the score at point $\boldsymbol{x}(t)$ at time $t$, i.e.,  $[s_1(\boldsymbol{\theta}),s_2(\boldsymbol{\theta})]=\boldsymbol{s}_{\boldsymbol{\theta}} =\boldsymbol{s}_{\boldsymbol{\theta}} (x_1(t),x_2(t),t,t)$.
 
As previously described, in SA-DPM the model attempts to learn all the scores of the evolution of the distribution dictated by the (DDPM) forward process: 
\begin{equation*}
d\boldsymbol{x}(t)=-\frac{1}{2}b(t)[\boldsymbol{x}(t)+\nabla_{\boldsymbol{x}(t)}\log{p_t(\boldsymbol{x}(t))}]dt;\ t\in [0,1],
\end{equation*} where we set $b(t)=10t$.

In contrast, for the TPSM$_A$ approach, we train two distinct models (networks), with the initial model spanning the time interval $[0,0.1]$ and the subsequent one covering $[0.1,1]$. In the case of TPSM$_B$, we partition the diffusion process into four subintervals with the following splits $\{0$, $0.02$, $0.1$, $0.3$, $1\}$. Conversely, in the extreme TPSM$_C$ approach, we utilize 200 such networks, where network $i$ is trained to learn the scores corresponding to the evolution of the distribution dictated by the identical process, constrained to the time interval $\ t\in [\frac{i}{200},\frac{i+1}{200}]$.

\begin{table}[b]
    \caption{A comparison of the results between SA-DPM and TPSM variants. The results are given in NLL (lower is better). Parallel training time is given in parentheses (blocks $\times$ hours per block).}
    \label{table2}
    \centering
    \begin{tabular}{lllll}
        \toprule
        Method: & SA-DPM & TPSM$_A$ & TPSM$_B$ & TPSM$_C$  \\
        \midrule
        \makecell{TY}     & 0.76 (1$\times$ \textbf{4h}) & 0.69 (2$\times$ \textbf{1h}) & 0.64 (4$\times$ \textbf{1h}) & 0.58 (200$\times$ \textbf{1h}) \\
        \makecell{HG}     & 1.11 (1$\times$ \textbf{1.3h}) & 1.07 (2$\times$ \textbf{0.3h}) & 1.04 (4$\times$ \textbf{0.3h}) & 1.03 (200$\times$ \textbf{0.3h}) \\
        \bottomrule
    \end{tabular}
    %\vspace{-10pt}
\end{table}

As evidenced in Figure \ref{toy2dimage}, TPSM demonstrates a substantially enhanced performance in the evaluated tasks, when juxtaposed with the baseline (SA-DPM), notwithstanding the fact that the latter underwent a training duration four times lengthier.
In the initial row, the intended target is the TY distribution, a notably intricate and arduous 2D distribution to model. It becomes apparent that the baseline exhibits difficulty in modeling such a distribution, as it fails to adequately delineate the distinct modalities constituting the probability density function. One could conjecture that a considerable enhancement of the model size would likely yield superior performance from the SA-DPM, albeit this would inevitably augment the already substantial training duration. Moreover, when considering more realistic scenarios involving high-dimensional data, the size of the network would inevitably be constrained due to memory-related limitations.
Analogous distinctions can be discerned in the second row (HG distribution). In Table \ref{table2}, we present the results of the test NLL for both models across both datasets, in addition to the difference in training time. The experiments demonstrated herein indicate that the TPSM approach is characterized by high flexibility of the vector field.

TPSM naturally retains the ability to generate samples from the learnt distribution and perform likelihood estimation by employing Equations \ref{g} and \ref{l} (Appendix \ref{AppendixA}). In addition, any ODE solver can be used, including adaptive ones, identically as in the case of CNFs. In our experiments we use the Runge-Kutta-4 (RK4) method and we precisely calculate the divergence during validation, as there is no necessity to utilize the Hutchinson's trace estimator \citep{doi:10.1080/03610919008812866, https://doi.org/10.48550/arxiv.1810.01367} due to the data's low dimensionality.

\subsection{Image Datasets}

In this section, we showcase the enhanced performance attained by employing PSM methods on image data. We parameterize the SA-DPM model using the time-varying U-Net architecture introduced in \citep{DBLP:journals/corr/abs-2006-11239} and set the number of channels to 64. The same network is utilized in each block for the TPSM approach.

The initial TPSM variant explored, designated as TPSM0, follows a similar approach as TPSM$_A$ in the synthetic 2D data case, wherein the diffusion interval $[0, 1]$ is partitioned into two subintervals: $[0, 0.1]$ and $[0.1, 1]$. By training each network in each block for half the number of parameter updates compared to the SA-DPM network, the training time is reduced by 50\% through parallelization. Moreover, improvements in likelihood estimation results are observed (Table \ref{tableTP0}), with no adverse effects on inference time or memory usage, as illustrated in Tables \ref{table3ima}, \ref{table3cel}, and \ref{table3cif} in Appendix \ref{AppendixB}. During the inference phase, 100 steps were executed within the interval $[0, 0.1]$ and 900 steps within $[0.1, 1]$.

\begin{table}[ht]
    \caption{The results of TPSM0, where the two time subintervals have unequal length. In parentheses we give the number of GPUs (blocks) $\times$ the time of training per block. In square brackets we give the number of GPUs (blocks) $\times$ the number of parameter updates per block.}
    \label{tableTP0}
    \centering
    \begin{tabular}{llll}
    \toprule
    Method: & CIFAR-10 & CelebA 64$\times$64 & ImageNet 64$\times$64\\
    \midrule
    SA-DPM  & 3.13 (1$\times$ \textbf{72h}) [1$\times$\textbf{800k}]  & 2.06 (1$\times$ \textbf{72h}) [1$\times$\textbf{800k}] & 3.62 (1$\times$ \textbf{180h}) [1$\times$\textbf{2000k}]\\
    TPSM0 & 3.12 (2$\times$ \textbf{36h}) [2$\times$\textbf{400k}]  & 1.92(2$\times$ \textbf{36h}) [2$\times$\textbf{400k}] & 3.55 (2$\times$ \textbf{90h}) [2$\times$\textbf{1000k}]\\
    \bottomrule
    \end{tabular}
\end{table}

Moreover, we evaluate TPSM1, which consists of 10 blocks, and to investigate the limiting behavior of TPSM, we introduce TPSM2, comprised of 100 blocks. For TPSM1, block $i$ models the score associated with times $\ t\in [\frac{i}{10},\frac{i+1}{10}]$, whereas in the case of TPSM2, this changes to $\ t\in [\frac{i}{100},\frac{i+1}{100}]$. The results are given in Table \ref{table3} and demonstrate significant improvements in density estimation and training time through parallelization. 

\begin{table}[b]
    \caption{Results comparing the performance of SA-DPM and the parallel score-matching approaches. We test the models on CIFAR-10, CelebA, and ImageNet (64x64). The results are given in bits/dim (lower is better), and the training time is given in parentheses.}
    \label{table3}
    \centering
    \begin{tabular}{llll}
    \toprule
    Method: & CIFAR-10 & CelebA 64$\times$64 & ImageNet 64$\times$64 \\
    \midrule
    SA-DPM & 3.13 (1$\times$ \textbf{72h})  & 2.06 (1$\times$ \textbf{72h}) & 3.62 (1$\times$ \textbf{180h})\\
    TPSM1 & 3.11 (10$\times$ \textbf{9h}) & 2.07 (10$\times$ \textbf{9h}) & 3.60 (10$\times$ \textbf{22h})\\
    TPSM2 & 2.93 (100$\times$ \textbf{4.5h})& 1.90 (100$\times$ \textbf{4.5h}) & 3.55 (100$\times$ \textbf{14h}) \\
    DPSM & 2.93 (1000$\times$ \textbf{1.5h}) & 1.94 (1000$\times$ \textbf{2.5h}) & 3.59 (1000$\times$ \textbf{7h}) \\
    \bottomrule
    \end{tabular}

\end{table}
We observe that TPSM0 outperforms TPSM1, attributable to the decision to utilize the DDPM setting, where the majority of the local score evolution transpires when $t$ approaches 0, \cite{pmlr-v139-nichol21a}. This insinuates that the results of TPSM1 and TPSM2 could experience significant enhancements if more recent and efficient settings were employed, such as Flow-Matching, \cite{lipman2023flow}, where the score's evolution is distributed more uniformly over time. Such is the case for 2D toy data as shown in Appendix \ref{AppendixF}.

\begin{figure}[H]
\centering
\begin{tabular}{ll}
\includegraphics[scale=0.605]{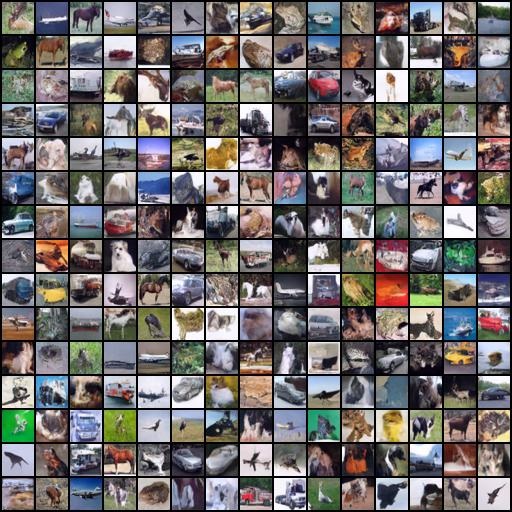}
&
\includegraphics[scale=0.28]{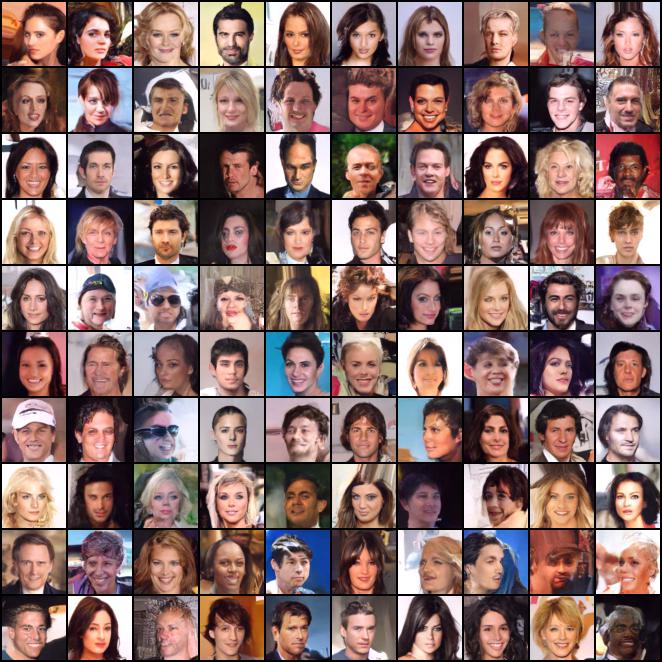}
\end{tabular}
\caption{Random non cherry-picked CIFAR-10 and CelebA samples generated by the DPSM with 1000 training/generative steps.}
\label{samplesddpmm}
\end{figure}
  As in the case of 2D data, we set $b(t)=10t$ in all cases. The RK4 solver (1k steps) was used during density estimation in SA-DPM, and in TPSM. If the Euler method is used, a higher number of integration steps is suggested ($\geq5$k), as otherwise sub-optimal results can be obtained,  which overestimate the performance of the model. Generated samples from such models can be found in Appendix \ref{AppendixC}. In Appendix \ref{AppendixB}, we provide results in the case that SA-DPM and TPSM2 are trained for the same number of parameter updates.

The model implemented in the DPSM framework differs significantly, as it does not depend on time. In this instance, we elect to employ the initial U-Net as introduced and implemented in the pioneering U-Net paper \citep{DBLP:journals/corr/RonnebergerFB15}. The channel architecture is configured as follows: $(3,a,a*2,a*4,a*8,a*16,a*8,a*4,a*2,a,3)$.

When $a$ is set to 64, this architecture aligns with the standard implementation. The discretization process incorporates 1000 steps, signifying the use of 1000 basic U-Nets, each per step. The parameters $b(t)$ are defined as previously,  $b(t)=10t$. The classical non time-varying U-Net exhibits approximately 3.5 times faster training time per parameter update compared to the time-varying U-Net. Coupled with fewer parameter updates, this accelerates DPSM training, making it up to 50 times quicker than that of SA-DPM. Additionally, the U-Net's reduced complexity facilitates usage of batch sizes that are four times larger than in the time-varying scenario. However, DPSM implementation necessitates interpolation techniques for precise likelihood estimation. For our experiments involving a thousand networks, we simply interpret the ODE as a piece-wise constant function, specifically, in CNF block $i$, the vector field is defined to be constant, and is described by the optimized score approximator $s_i(\boldsymbol{\theta})$ at time $t_i$. Utilizing the Euler method, likelihood estimation is performed with 5k steps and generation with 1k steps, avoiding interpolation in the generation process. Figure \ref{samplesddpmm} presents generated samples of CIFAR-10 and CelebA. Detailed information on the number of parameter updates for each model and dataset is available in Table \ref{table4} in Appendix \ref{AppendixB}. Although the main focus of this paper is on density estimation performance quantified using rigorous scientific metrics like KL divergence (bits/dim), in Appendix \ref{AppendixB}, we present results relating to the quality of generated samples, evaluated using the widely accepted 2-Wasserstein distance approximation (FID), \cite{Seitzer2020FID}.

\section{Limitations and Future Work}

\subsection{Number of Computing Units and Likelihood Estimation in DPSM}
The DPSM framework fundamentally relies on parallel computing, which necessitates a substantial number of computing units. In fact, the true advantages of parallelization within DPSM can only be fully exploited when one has access to computing clusters. Conversely, the TPSM framework can be effectively operated even by those with access to a single GPU, provided that the number of blocks is kept low. As demonstrated earlier, training two CNF blocks sequentially (as in the TPSM0 scenario) requires the same duration as training a single block within the SA-DPM framework. However, this approach results in enhanced performance, as evidenced in Table \ref{tableTP0}.

Though the training process can be performed in parallel, the generation and likelihood estimation processes are inherently sequential. In the context of DPSM, it is necessary to interpolate between successive models to carry out accurate likelihood estimation. The generation process, on the other hand, remains unaffected.

\subsection{Tailored TPSM Models}

In this study, due to computational simplifications, we chose to employ smaller, older models across all three frameworks (SA-DPM, TPSM, DPSM). Our main objective was to demonstrate that the latter two methods outperform the first one. In all instances, for fair comparison, we used models in the individual blocks (steps) in the TPSM (DPSM) approach that were equal to or smaller than the ones used in SA-DPM. Looking ahead, it would be valuable to remove these restrictions and attempt to develop custom models for the TPSM/DPSM frameworks. This would allow us to assess how much these strategies can enhance the state of the art.

\section{Conclusion}

We have presented Parallel Score Matching strategies for training diffusion probabilistic models. We exploited the inherent properties of diffusion models which enable modeling of each score separately. We showed that learning different groups of scores in parallel via independent neural networks is effective and allows great improvements of  training time, while enabling better model performance. 

\newpage
\bibliographystyle{apa-good} % Style BST file (imsart-number.bst or imsart-nameyear.bst)
\bibliography{main} 

\begin{thebibliography}{26}
\expandafter\ifx\csname natexlab\endcsname\relax\def\natexlab#1{#1}\fi
\expandafter\ifx\csname url\endcsname\relax
  \def\url#1{{\tt #1}}\fi
\expandafter\ifx\csname urlprefix\endcsname\relax\def\urlprefix{URL }\fi

\bibitem[{Behrmann et~al.(2019)Behrmann, Grathwohl, Chen, Duvenaud, \&
  Jacobsen}]{pmlr-v97-behrmann19a}
Behrmann, J., Grathwohl, W., Chen, R. T.~Q., Duvenaud, D., \& Jacobsen, J.-H.
  (2019).
\newblock Invertible residual networks.
\newblock In K.~Chaudhuri, \& R.~Salakhutdinov (Eds.) {\em Proceedings of the
  36th International Conference on Machine Learning\/}, vol.~97 of {\em
  Proceedings of Machine Learning Research\/}, (pp. 573--582). PMLR.

\bibitem[{Chen et~al.(2018)Chen, Rubanova, Bettencourt, \&
  Duvenaud}]{https://doi.org/10.48550/arxiv.1806.07366}
Chen, R. T.~Q., Rubanova, Y., Bettencourt, J., \& Duvenaud, D.~K. (2018).
\newblock Neural ordinary differential equations.
\newblock In S.~Bengio, H.~Wallach, H.~Larochelle, K.~Grauman, N.~Cesa-Bianchi,
  \& R.~Garnett (Eds.) {\em Advances in Neural Information Processing
  Systems\/}, vol.~31. Curran Associates, Inc.

\bibitem[{Deng et~al.(2009)Deng, Dong, Socher, Li, Li, \&
  Fei-Fei}]{deng2009imagenet}
Deng, J., Dong, W., Socher, R., Li, L.-J., Li, K., \& Fei-Fei, L. (2009).
\newblock Imagenet: A large-scale hierarchical image database.
\newblock In {\em 2009 IEEE conference on computer vision and pattern
  recognition\/}, (pp. 248--255). Ieee.

\bibitem[{Dinh et~al.(2015)Dinh, Krueger, \&
  Bengio}]{https://doi.org/10.48550/arxiv.1410.8516}
Dinh, L., Krueger, D., \& Bengio, Y. (2015).
\newblock {NICE:} non-linear independent components estimation.
\newblock In Y.~Bengio, \& Y.~LeCun (Eds.) {\em 3rd International Conference on
  Learning Representations, {ICLR} 2015, San Diego, CA, USA, May 7-9, 2015,
  Workshop Track Proceedings\/}.

\bibitem[{Grathwohl et~al.(2019)Grathwohl, Chen, Bettencourt, \&
  Duvenaud}]{https://doi.org/10.48550/arxiv.1810.01367}
Grathwohl, W., Chen, R. T.~Q., Bettencourt, J., \& Duvenaud, D. (2019).
\newblock Scalable reversible generative models with free-form continuous
  dynamics.
\newblock In {\em International Conference on Learning Representations\/}.

\bibitem[{Ho et~al.(2020)Ho, Jain, \&
  Abbeel}]{DBLP:journals/corr/abs-2006-11239}
Ho, J., Jain, A., \& Abbeel, P. (2020).
\newblock Denoising diffusion probabilistic models.
\newblock In H.~Larochelle, M.~Ranzato, R.~Hadsell, M.~Balcan, \& H.~Lin (Eds.)
  {\em Advances in Neural Information Processing Systems\/}, vol.~33, (pp.
  6840--6851). Curran Associates, Inc.

\bibitem[{Hutchinson(1990)}]{doi:10.1080/03610919008812866}
Hutchinson, M. (1990).
\newblock A stochastic estimator of the trace of the influence matrix for
  laplacian smoothing splines.
\newblock {\em Communications in Statistics - Simulation and Computation\/},
  {\em 19\/}(2), 433--450.

\bibitem[{Hyv{{\"a}}rinen(2005)}]{JMLR:v6:hyvarinen05a}
Hyv{{\"a}}rinen, A. (2005).
\newblock Estimation of non-normalized statistical models by score matching.
\newblock {\em Journal of Machine Learning Research\/}, {\em 6\/}(24),
  695--709.

\bibitem[{Kingma et~al.(2021)Kingma, Salimans, Poole, \&
  Ho}]{DBLP:journals/corr/abs-2107-00630}
Kingma, D.~P., Salimans, T., Poole, B., \& Ho, J. (2021).
\newblock On density estimation with diffusion models.
\newblock In A.~Beygelzimer, Y.~Dauphin, P.~Liang, \& J.~W. Vaughan (Eds.) {\em
  Advances in Neural Information Processing Systems\/}.

\bibitem[{Kobyzev et~al.(2021)Kobyzev, Prince, \& Brubaker}]{Kobyzev_2021}
Kobyzev, I., Prince, S.~J., \& Brubaker, M.~A. (2021).
\newblock Normalizing flows: An introduction and review of current methods.
\newblock {\em IEEE Transactions on Pattern Analysis and Machine
  Intelligence\/}, {\em 43\/}(11), 3964--3979.

\bibitem[{Kynk{\"a}{\"a}nniemi et~al.(2023)Kynk{\"a}{\"a}nniemi, Karras,
  Aittala, Aila, \& Lehtinen}]{pseudoscientificmetric}
Kynk{\"a}{\"a}nniemi, T., Karras, T., Aittala, M., Aila, T., \& Lehtinen, J.
  (2023).
\newblock The role of imagenet classes in fr\'echet inception distance.
\newblock In {\em The Eleventh International Conference on Learning
  Representations\/}.

\bibitem[{Lipman et~al.(2023)Lipman, Chen, Ben-Hamu, Nickel, \&
  Le}]{lipman2023flow}
Lipman, Y., Chen, R. T.~Q., Ben-Hamu, H., Nickel, M., \& Le, M. (2023).
\newblock Flow matching for generative modeling.
\newblock In {\em The Eleventh International Conference on Learning
  Representations\/}.

\bibitem[{Liu et~al.(2015)Liu, Luo, Wang, \& Tang}]{liu2015faceattributes}
Liu, Z., Luo, P., Wang, X., \& Tang, X. (2015).
\newblock Deep learning face attributes in the wild.
\newblock In {\em Proceedings of International Conference on Computer Vision
  (ICCV)\/}.

\bibitem[{Nichol \& Dhariwal(2021)}]{pmlr-v139-nichol21a}
Nichol, A.~Q., \& Dhariwal, P. (2021).
\newblock Improved denoising diffusion probabilistic models.
\newblock In M.~Meila, \& T.~Zhang (Eds.) {\em Proceedings of the 38th
  International Conference on Machine Learning\/}, vol. 139 of {\em Proceedings
  of Machine Learning Research\/}, (pp. 8162--8171). PMLR.

\bibitem[{Papamakarios et~al.(2021)Papamakarios, Nalisnick, Rezende, Mohamed,
  \& Lakshminarayanan}]{JMLR:v22:19-1028}
Papamakarios, G., Nalisnick, E., Rezende, D.~J., Mohamed, S., \&
  Lakshminarayanan, B. (2021).
\newblock Normalizing flows for probabilistic modeling and inference.
\newblock {\em Journal of Machine Learning Research\/}, {\em 22\/}(57), 1--64.

\bibitem[{Rezende \& Mohamed(2015)}]{https://doi.org/10.48550/arxiv.1505.05770}
Rezende, D., \& Mohamed, S. (2015).
\newblock Variational inference with normalizing flows.
\newblock In F.~Bach, \& D.~Blei (Eds.) {\em Proceedings of the 32nd
  International Conference on Machine Learning\/}, vol.~37 of {\em Proceedings
  of Machine Learning Research\/}, (pp. 1530--1538). Lille, France: PMLR.

\bibitem[{Rombach et~al.(2022)Rombach, Blattmann, Lorenz, Esser, \&
  Ommer}]{DBLP:journals/corr/abs-2112-10752}
Rombach, R., Blattmann, A., Lorenz, D., Esser, P., \& Ommer, B. (2022).
\newblock High-resolution image synthesis with latent diffusion models.
\newblock In {\em Proceedings of the IEEE/CVF Conference on Computer Vision and
  Pattern Recognition (CVPR)\/}, (pp. 10684--10695).

\bibitem[{Ronneberger et~al.(2015)Ronneberger, Fischer, \&
  Brox}]{DBLP:journals/corr/RonnebergerFB15}
Ronneberger, O., Fischer, P., \& Brox, T. (2015).
\newblock U-net: Convolutional networks for biomedical image segmentation.
\newblock In N.~Navab, J.~Hornegger, W.~M. Wells, \& A.~F. Frangi (Eds.) {\em
  Medical Image Computing and Computer-Assisted Intervention -- MICCAI 2015\/},
  (pp. 234--241). Cham: Springer International Publishing.

\bibitem[{Seitzer(2020)}]{Seitzer2020FID}
Seitzer, M. (2020).
\newblock {pytorch-fid: FID Score for PyTorch}.
\newblock Version 0.3.0.

\bibitem[{Sohl-Dickstein et~al.(2015)Sohl-Dickstein, Weiss, Maheswaranathan, \&
  Ganguli}]{DBLP:journals/corr/Sohl-DicksteinW15}
Sohl-Dickstein, J., Weiss, E., Maheswaranathan, N., \& Ganguli, S. (2015).
\newblock Deep unsupervised learning using nonequilibrium thermodynamics.
\newblock In F.~Bach, \& D.~Blei (Eds.) {\em Proceedings of the 32nd
  International Conference on Machine Learning\/}, vol.~37 of {\em Proceedings
  of Machine Learning Research\/}, (pp. 2256--2265). Lille, France: PMLR.

\bibitem[{Song et~al.(2020)Song, Garg, Shi, \&
  Ermon}]{DBLP:journals/corr/abs-1905-07088}
Song, Y., Garg, S., Shi, J., \& Ermon, S. (2020).
\newblock Sliced score matching: A scalable approach to density and score
  estimation.
\newblock In R.~P. Adams, \& V.~Gogate (Eds.) {\em Proceedings of The 35th
  Uncertainty in Artificial Intelligence Conference\/}, vol. 115 of {\em
  Proceedings of Machine Learning Research\/}, (pp. 574--584). PMLR.

\bibitem[{Song et~al.(2021)Song, Sohl-Dickstein, Kingma, Kumar, Ermon, \&
  Poole}]{song2021scorebased}
Song, Y., Sohl-Dickstein, J., Kingma, D.~P., Kumar, A., Ermon, S., \& Poole, B.
  (2021).
\newblock Score-based generative modeling through stochastic differential
  equations.
\newblock In {\em International Conference on Learning Representations\/}.

\bibitem[{Tabak \& Turner(2013)}]{https://doi.org/10.1002/cpa.21423}
Tabak, E.~G., \& Turner, C.~V. (2013).
\newblock A family of nonparametric density estimation algorithms.
\newblock {\em Communications on Pure and Applied Mathematics\/}, {\em
  66\/}(2), 145--164.

\bibitem[{Tabak \& Vanden-Eijnden(2010)}]{cms/1266935020}
Tabak, E.~G., \& Vanden-Eijnden, E. (2010).
\newblock {Density estimation by dual ascent of the log-likelihood}.
\newblock {\em Communications in Mathematical Sciences\/}, {\em 8\/}(1), 217 --
  233.

\bibitem[{Vincent(2011)}]{6795935}
Vincent, P. (2011).
\newblock A connection between score matching and denoising autoencoders.
\newblock {\em Neural Computation\/}, {\em 23\/}(7), 1661--1674.

\bibitem[{Wang(2020)}]{ddpmtorch}
Wang, P. (2020).
\newblock {https://github.com/lucidrains/denoising-diffusion-pytorch}.

\end{thebibliography}

\newpage
\appendix
\onecolumn

\section{Image Generation and Likelihood Estimation in TPSM and DPSM}\label{AppendixA}
Regarding the time-varying Parallel Score Matching (TPSM) approach, we can generate data as in the original framework by iterating the following integration process:
\begin{equation}\label{g}
\boldsymbol{x}(t_{i})=\boldsymbol{x}(t_{i+1})+\int_{t_i+1}^{t_{i}} \big[-\frac{1}{2}b(\tau)\big(\boldsymbol{x}(\tau)+s_{\boldsymbol{\theta}_{i}} (\boldsymbol{x}_{\tau}, \tau)\big)\big]d\tau, 
\end{equation}
and perform likelihood estimation via
\begin{equation*}
    \log{p(\boldsymbol{x}(t_0))}=\log{p(\boldsymbol{x}(t_N))}
\end{equation*}
\begin{equation}\label{l}
   +\Sigma_{i=0}^{N-1} \int_{t_i}^{t_{i+1}} \big[-\frac{1}{2}b(\tau) tr{\frac{\partial  \big(\boldsymbol{x}(\tau)+s_{\boldsymbol{\theta}_{i}} (\boldsymbol{x}_{\tau}, \tau)\big)}{\partial \boldsymbol{x}(\tau)}\big]d\tau},
\end{equation}
where $\boldsymbol{s}_{\boldsymbol{\theta}_{i}} (\boldsymbol{x}_{t}, t)=-\frac{\boldsymbol{\varepsilon}_{\boldsymbol{\theta}_{i}} (\boldsymbol{x}_{t}, t)}{\sqrt{1-e^{-\int_0^t b(s)ds}}}$.

 Similarly, in the case of Discrete Parallel Score Matching (DPSM), we can easily generate data via back-integrating
\begin{equation*}
\boldsymbol{x}(t_i)=\boldsymbol{x}(t_{i+1})-\epsilon \big[-\frac{1}{2}b(t_{i+1})\big(\boldsymbol{x}(t_{i+1})+s_{\boldsymbol{\theta}_{t_{i+1}}} (\boldsymbol{x}_{t_{i+1}})\big)\big]
\end{equation*}
Likewise, we can perform likelihood estimation as follows
\begin{equation*}
    \log{p(\boldsymbol{x}(t_0))}=\log{p(\boldsymbol{x}(t_N))}
\end{equation*}
\begin{equation*}
   +\Sigma_{i=1}^{N}\big[-\epsilon \frac{1}{2}b(t_i) tr{\frac{\partial  \big(\boldsymbol{x}(t_i)+s_{\boldsymbol{\theta}_{i}} (\boldsymbol{x}_{t_i})\big)}{\partial \boldsymbol{x}(t_i)}}\big].
\end{equation*}

In addition, in both approaches data samples can be generated by integrating the corresponding reverse SDE \citep{song2021scorebased}. 
\begin{equation*}
d\boldsymbol{x}(t)=-b(t)[\frac{1}{2}\boldsymbol{x}(t)+\nabla_{\boldsymbol{x}(t)}\log{p_t(\boldsymbol{x}(t))}]dt+ \sqrt{b(t)} d\boldsymbol{w}.
\end{equation*}

\section{Additional Results}\label{AppendixB}
Below we give additional results with regards to the experimental section.

First, we set up a comparison between TPSM2 and SA-DPM, where the number of parameter updates for each network is the same in both approaches. To elaborate, each network in TPSM2 undergoes 50k parallel parameter updates, and the network in SA-DPM also experiences 50k parameter updates. Given this configuration, the training duration for both approaches is equivalent. However, as displayed in Table \ref{table_appendix}, the disparity in their performance is quite significant.
\begin{table}[ht]
    \caption{Results comparing the performance of SA-DPM and TPSM2 for the same number of parameter updates. We test the models on CIFAR-10, CelebA, and ImageNet. The results are given in bits/dim (lower is better), the number of parameter updates is given in parentheses, and the training time is given in square brackets. }
    \label{table_appendix}
    \centering
    \begin{tabular}{llll}
    \toprule
    Method: & CIFAR-10 & CelebA 64$\times$64 & ImageNet 64$\times$64\\
    \midrule
    SA-DPM     & 3.30 (50k) [4.5H]  & 2.38 (50k) [4.5H] & 3.76 (150k) [14H]\\
    TPSM2     & 2.93 (50k) [4.5H] & 1.90 (50k) [4.5H] & 3.55 (150k) [14H]\\
    \bottomrule
    \end{tabular}
\vskip -0.1in
\end{table}

% Then in Table \ref{table4}, we provide the number of parameter updates during training in the different approaches compared in  \ref{table3}, in the main text of the paper. 

\begin{table}[H]
    \caption{Results related to Table \ref{table3}, comparing number of parameter updates of SA-DPMs and the parallel score matching approaches, namely TPSM and DPSM. }
    \label{table4}
    \centering
    \begin{tabular}{llll}
    \toprule
    Method: & CIFAR-10 & CelebA & ImageNet\\
    \midrule
    SA-DPM & 800k  & 800k & 2000k \\
    
    TPSM0 & 400k & 400k  & 1000k \\
    
    TPSM1 & 100k & 100k  & 250k \\
    
    TPSM2 & 50k & 50k & 150k \\
    
    DPSM & 50k & 100k  & 250k \\
    \bottomrule
    \end{tabular}

\end{table}

\begin{figure}[ht]
\centering
\begin{tabular}{l}
\includegraphics[scale=0.5]{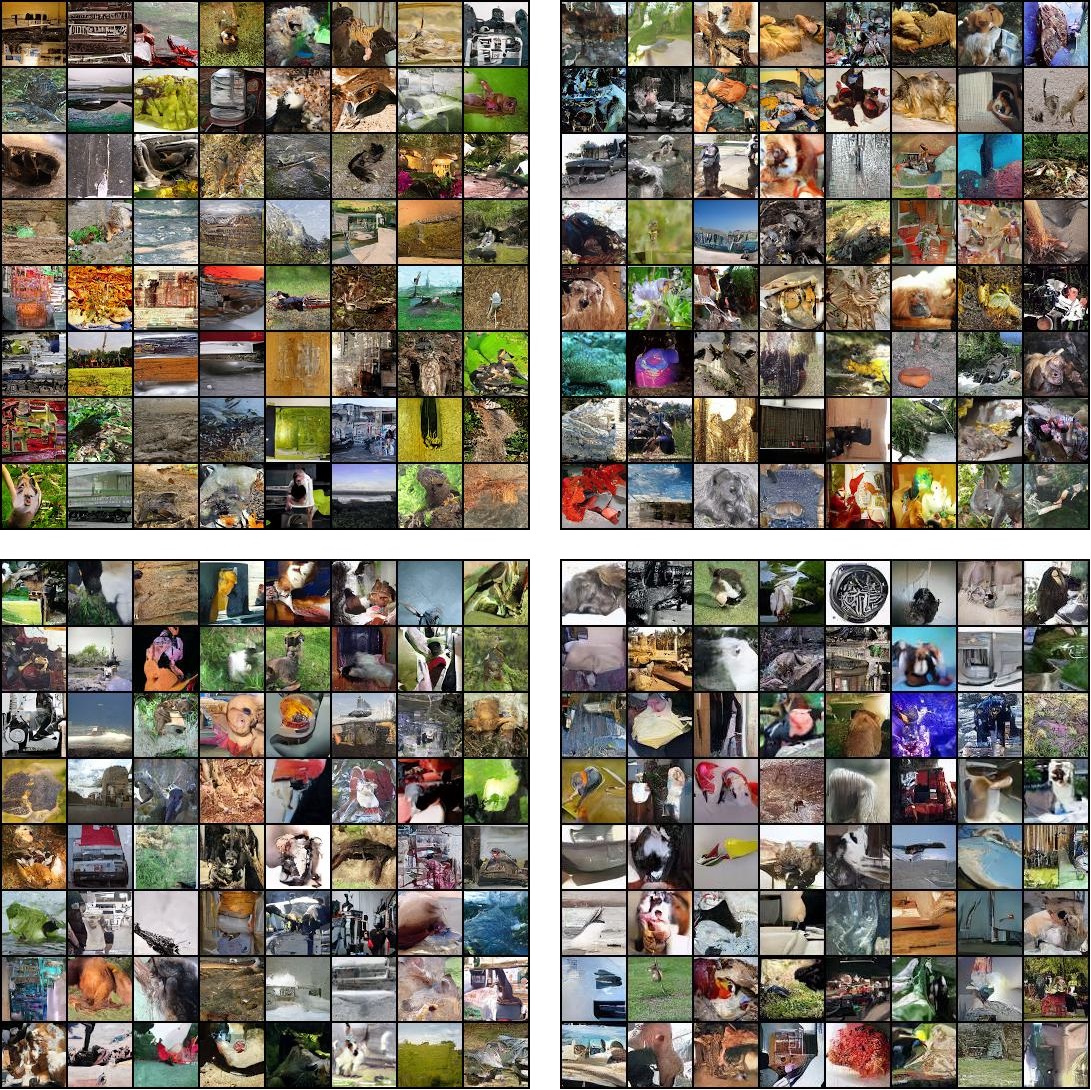}
\end{tabular}
\caption{Random non cherry-picked generated ImageNet 64x64 images. Top-Left: SA-DPM, 2000k parameter updates, 1 CNF block, 10k sampling steps via the reverse SDE. Top-Right: TPSM, 250k parameter updates, 10 CNF blocks, 10k sampling steps via the reverse SDE. Bottom-Left: TPSM, 150k parameter updates, 100 CNF blocks, 10k sampling steps via the reverse SDE. Bottom-Right: DPSM, 250k parameter updates, 1000 CNF blocks, 1k sampling steps via the reverse SDE.}
\label{generated_imagenet}
\end{figure}

\begin{figure}[H]
\centering
\begin{tabular}{l}
\includegraphics[scale=0.8]{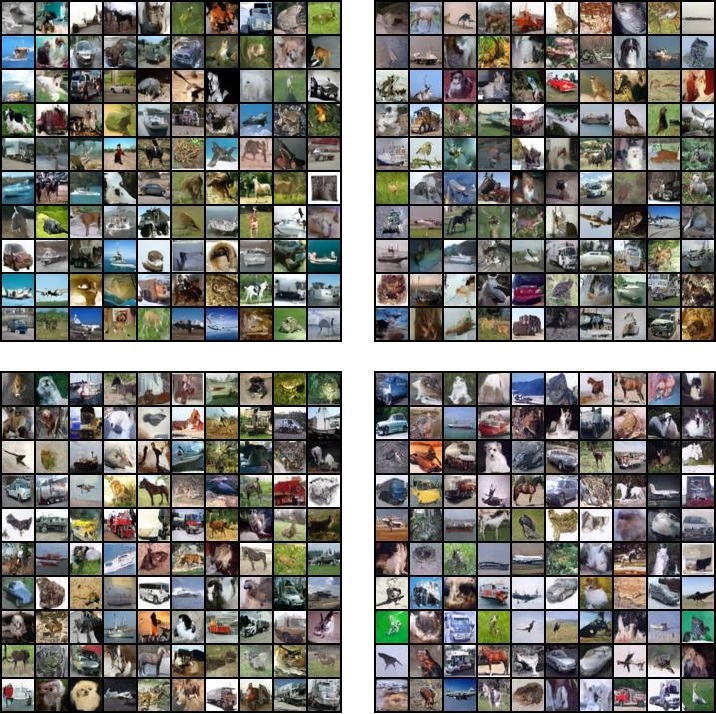}
\end{tabular}
\caption{Random non cherry-picked generated CIFAR-10 images. Top-Left: SA-DPM, 800k parameter updates, 1 CNF block, 10k sampling steps via the reverse SDE. Top-Right: TPSM, 100k parameter updates, 10 CNF blocks, 10k sampling steps via the reverse SDE. Bottom-Left: TPSM, 50k parameter updates, 100 CNF blocks, 10k sampling steps via the reverse SDE. Bottom-Right: DPSM, 50k parameter updates, 1000 CNF blocks, 1k sampling steps via the reverse SDE.}
\label{generated_cifar}
\end{figure}

In addition, below, we provide results as measured by the usual approximation of FID. The number of parameter updates is identical as before (Table \ref{table4}). In all cases we generated 50 thousand samples. It is important to notice that while for example the images generated by TPSM2 in Figures \ref{generated_imagenet} and \ref{generated_cifar} are visibly better than those by SA-DPM, the FID score does not reflect this, likely due to it being limited by the several underlying  assumptions and biases inflicted by arbitrary choices in its definition \cite{pseudoscientificmetric}. 
\begin{table}[H]
    \caption{The results given in FID.}
    \label{table_fid}
    \centering
    \begin{tabular}{llll}
    \toprule
    Method: & CIFAR-10 & CelebA & ImageNet\\
    \midrule
    SA-DPM (3k SDE steps)    & 12.45 & 5.34 & 48.8  \\
    TPSM0 (3k SDE steps)    & 12.88  & 8.19  & 51.2 \\
    TPSM1 (3k SDE steps)    & 9.54  & 7.97 & 51.9 \\
    TPSM2 (3k SDE steps)    & 12.83 & 9.23  & 48.7  \\
    SA-DPM (1k SDE steps)    & 12.47  & 5.40  & 51.5  \\
    DPSM  (1k SDE steps)   & \textbf{8.89}  & \textbf{4.88}  & \textbf{38.06}  \\
    \bottomrule
    \end{tabular}
\end{table}

Furthermore, we present inferential outcomes in relation to generation and likelihood estimation time, as well as the memory utilization of all the considered approaches. It is noteworthy that PSM approaches not only abstain from inducing inferential penalties, but also have the potential to expedite generation and likelihood estimation in certain cases, such as DPSM. The batch size for SDE and ODE generation was consistently maintained at 32, whereas for likelihood estimation, it was uniformly reduced to 16. In all instances, the generation of SDEs was executed employing 1000 Numerical Function Evaluations (NFE). In contrast, for ODE generation and likelihood estimation pertaining to SA-DPM and TPSM, a total of 1000 steps employing the RK4 method were taken, equating to 4000 function evaluations. Specifically, in the context of DPSM, a total of 4000 steps were performed using interpolation and integration via the Euler method for both ODE generation and likelihood estimation. The results are provided in Tables \ref{table3ima}, \ref{table3cel} and \ref{table3cif}.
\begin{table}[H]
    \caption{Inference results on ImageNet. Inference time is given in seconds while memory usage is provided in MB.}
    \label{table3ima}
    \centering
    \begin{tabular}{llll}
    \toprule
    Task & Generation SDE & Generation ODE & Likelihood Estimation\\
    \midrule
    SA-DPM & 131.8s (2396MB) & 674.4s (4260MB)  & 858.8s (7164MB)  \\
    TPSM0 & 119.8s (2334MB)  & 635.7s (3951MB) &  936.2s (6221MB) \\ 
    TPSM1 & 123.6s (2396MB)  & 626.8s (2800MB) & 887.7s (6624MB) \\
    TPSM2 & 137.1s (2381MB) & 567.3s (2392MB) & 792.4s (6187MB)\\
    DPSM & 124.7s (2709MB) & 326.2s (2569MB) &  335.2s (2670MB)\\
    \bottomrule
\end{tabular}

%\vskip -0.1in
\end{table}

\begin{table}[H]
    \caption{Inference results on CelebA. Inference time is given in seconds while memory usage is provided in MB.}
    \label{table3cel}
    \centering
    \begin{tabular}{llll}
    \toprule
    Task & Generation SDE & Generation ODE & Likelihood Estimation\\
    \midrule
    SA-DPM & 121.2s (2450MB) & 662.8s (4156MB)  & 892.2s (7139MB)  \\
    TPSM0 & 116.3s (2402MB)  & 599.3s (3704MB) &  933.9s (6261MB) \\ 
    TPSM1 & 126.9s (2426MB)  & 661.4s (2789MB) & 929.3s (6470MB) \\
    TPSM2 & 146s (2435MB) & 605.5s (2401MB) & 837.8s (616.8MB)\\
    DPSM & 116.6s (2518MB) & 303.1s (2529MB) &  318.8s (2650MB)\\
    \bottomrule
    \end{tabular}

\end{table}

\begin{table}[H]
    \caption{Inference results on CIFAR-10. Inference time is given in seconds while memory usage is provided in MB.}
    \label{table3cif}
    \centering
    \begin{tabular}{llll}
    \toprule
    Task & Generation SDE & Generation ODE & Likelihood Estimation\\
    \midrule
    SA-DPM & 38.3s (1890MB) & 184.7s (2279MB)  & 380.2s (3528MB)  \\
    TPSM0 & 38.5s (1780MB)  & 195.2s (2252MB) &  354s (2906MB) \\ 
    TPSM1 & 37.9s (1900MB)  & 198.3s (1989MB) & 393.8s (2964MB) \\
    TPSM2 & 51.1s (1883MB) & 185.6s (1895MB) & 367s (2920MB)\\
    DPSM & 95.2s (1930MB) &  171.3s (1959MB) &  217.8s (1959MB)\\
    \bottomrule
    \end{tabular}

\end{table}

\section{Additional Generated Samples}\label{AppendixC}
Below we provide generated samples from all the models that are present in Table \ref{table3}.

\begin{figure}[H]
\centering
\begin{tabular}{l}
\includegraphics[scale=0.5]{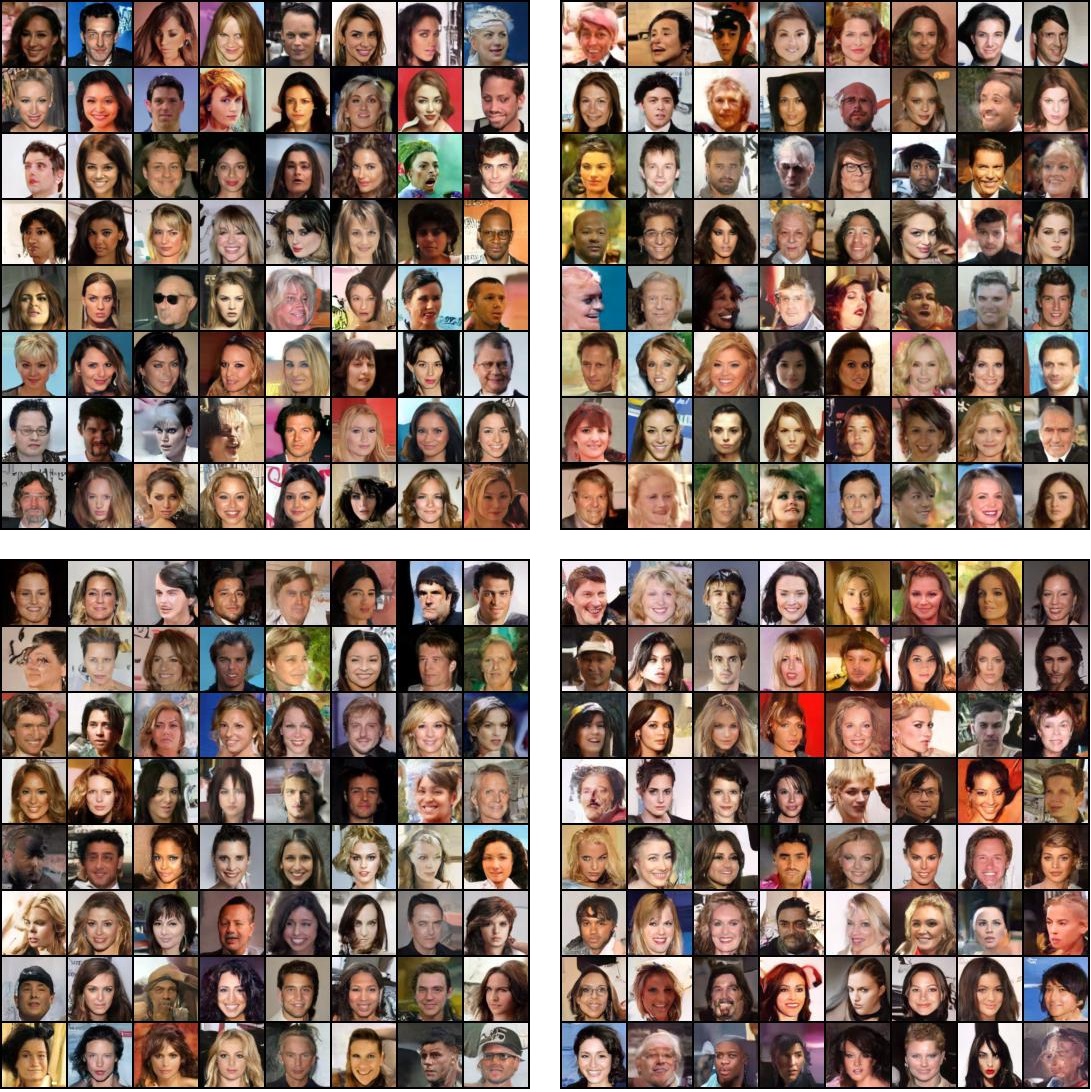}
\end{tabular}
\caption{Random non cherry-picked generated CelebA images. Top-Left: SA-DPM, 800k parameter updates, 1 CNF block, 10k sampling steps via the reverse SDE. Top-Right: TPSM, 100k parameter updates, 10 CNF blocks, 10k sampling steps via the reverse SDE. Bottom-Left: TPSM, 50k parameter updates, 100 CNF blocks, 10k sampling steps via the reverse SDE. Bottom-Right: DPSM, 100k parameter updates, 1000 CNF blocks, 1k sampling steps via the reverse SDE.}
\label{generated_celeba}
\end{figure}

\begin{figure}[H]
\centering
\begin{tabular}{l}
\includegraphics[scale=0.9]{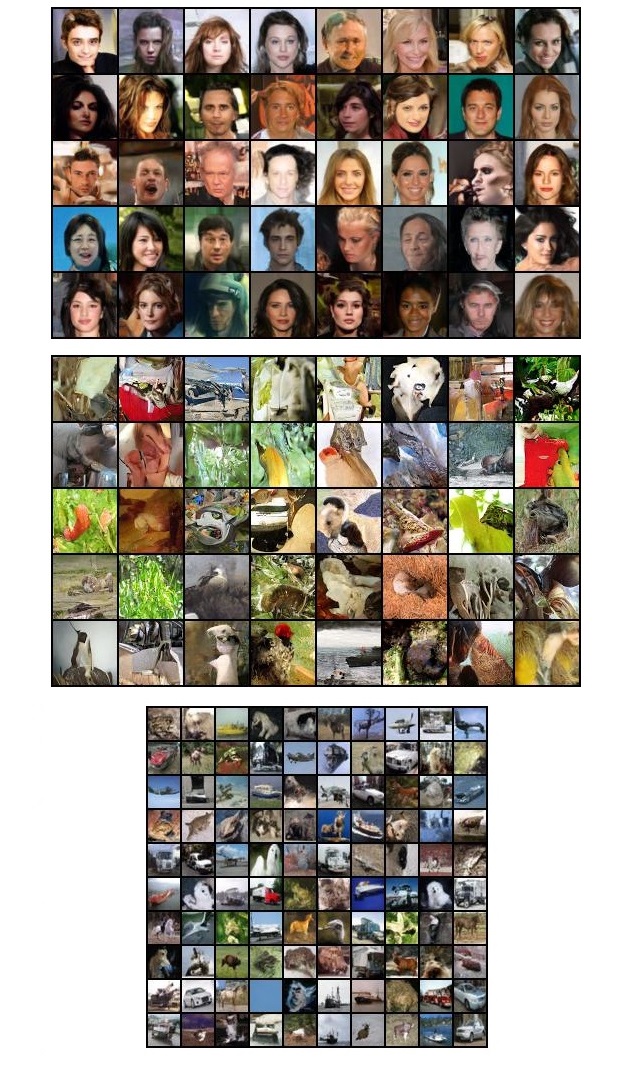}
\end{tabular}
\caption{Random non cherry-picked generated CelebA, ImageNet64x64 and Cifar-10 images generated by TPSM0 with 2 CNF blocks and 10000 reverse SDE steps. The number of parameter updates is half that of SA-DPM per block as given in Table \ref{table4}.}
\label{generated_tpsm0}
\end{figure}
\newpage

\section{The Continuous Evolution of the Score During Diffusion}\label{AppendixD}

The SDE process of transforming the data distribution to a normal one is the following:
\begin{equation}\label{1}
    d\boldsymbol{x}(t)=-\frac{1}{2}b(t)\boldsymbol{x}(t)dt+ \sqrt{b(t)} d\boldsymbol{w}.
\end{equation}

The equivalent process in terms of the evolution of the distribution is given by the following ODE:
\begin{equation}\label{2}
    d\boldsymbol{x}(t)=-\frac{1}{2}b(t)\big(\boldsymbol{x}(t) + \nabla_{\boldsymbol{x}(t)}\log p(\boldsymbol{x}(t),t)\big)dt=f(\boldsymbol{x}(t),t)dt.
\end{equation}

The instantaneous change of variable formula states that,
\begin{equation}\label{3}
    \frac{d\log p(\boldsymbol{x}(t),t)}{dt}=-tr\frac{\partial f(\boldsymbol{x}(t),t)}{\partial \boldsymbol{x}(t)},
\end{equation}
thus by substituting $f$:
\begin{equation}\label{4}
    \frac{d\log p(\boldsymbol{x}(t),t)}{dt}=\frac{1}{2} b(t)tr\frac{\partial \boldsymbol{x}(t) + \nabla_{\boldsymbol{x}(t)}\log p(\boldsymbol{x}(t),t)}{\partial \boldsymbol{x}(t)}=\frac{1}{2}b(t)D+ \frac{1}{2}b(t) tr\frac{\partial \nabla_{\boldsymbol{x}(t)}\log p(\boldsymbol{x}(t),t)}{\partial \boldsymbol{x}(t)}.
\end{equation}
It is easy to notice that,
\begin{equation}\label{5}
    \frac{d \nabla_{\boldsymbol{x}(t)}  \log p(\boldsymbol{x}(t),t)}{dt}= \nabla_{\boldsymbol{x}(t)} \frac{d   \log p(\boldsymbol{x}(t),t)}{dt}=\frac{1}{2}b(t)\nabla_{\boldsymbol{x}(t)} tr\frac{\partial \nabla_{\boldsymbol{x}(t)}\log p(\boldsymbol{x}(t),t)}{\partial \boldsymbol{x}(t)}.
\end{equation}

If we denote $s(\boldsymbol{x}(t),t)=\nabla_{\boldsymbol{x}(t)}\log p(\boldsymbol{x}(t),t)$, the last expression becomes:

\begin{equation}\label{6}
    \frac{d s(\boldsymbol{x}(t),t)}{dt}= \frac{1}{2}b(t)\nabla_{\boldsymbol{x}(t)} tr\frac{\partial s(\boldsymbol{x}(t),t)}{\partial \boldsymbol{x}(t)}.
\end{equation}

Since, 

\begin{equation}\label{7}
    \frac{d s(\boldsymbol{x}(t),t)}{dt}= \frac{\partial s(\boldsymbol{x}(t),t)}{\partial \boldsymbol{x}(t)} \frac{d \boldsymbol{x}(t)}{dt}+ \frac{\partial s(\boldsymbol{x}(t),t)}{\partial t}
\end{equation}

using equation \ref{6}, we get:

\begin{equation}\label{8}
    \frac{\partial s(\boldsymbol{x}(t),t)}{\partial \boldsymbol{x}(t)} \frac{d \boldsymbol{x}(t)}{dt}+ \frac{\partial s(\boldsymbol{x}(t),t)}{\partial t} = \frac{1}{2}b(t)\nabla_{\boldsymbol{x}(t)} tr\frac{\partial s(\boldsymbol{x}(t),t)}{\partial \boldsymbol{x}(t)},
\end{equation}

and therefore

\begin{equation}\label{9}
     \frac{\partial s(\boldsymbol{x},t)}{\partial t} = \frac{1}{2}b(t)\nabla_{\boldsymbol{x}} tr\frac{\partial s(\boldsymbol{x},t)}{\partial \boldsymbol{x}}-\frac{\partial s(\boldsymbol{x},t)}{\partial \boldsymbol{x}} \frac{d \boldsymbol{x}}{dt}.
\end{equation}

We conclude that the score evolves continuously in time as described by the following PDE:

\begin{equation}\label{10}
     \frac{\partial s(\boldsymbol{x},t)}{\partial t} = \frac{1}{2}b(t)\nabla_{\boldsymbol{x}} tr\frac{\partial s(\boldsymbol{x},t)}{\partial \boldsymbol{x}}+\frac{1}{2}b(t)\frac{\partial s(\boldsymbol{x},t)}{\partial \boldsymbol{x}} (\boldsymbol{x}+s(\boldsymbol{x},t)).
\end{equation}
This emphasizes the fact that the scores of all the intermediate distributions defined by the FDP are completely determined by the score of the initial (data) distribution. 
\newpage

\section{Standard Approach for DPMs}\label{AppendixE}

\subsection{Diffusion Probabilistic Models (DPMs)}

\textbf{Denoising Diffusion Models:}
A forward diffusion process or diffusion process \citep{DBLP:journals/corr/Sohl-DicksteinW15} is a fixed Markov chain that
gradually adds Gaussian noise to the data according to a schedule $\{b_t|t \in [n] \}$:
\begin{equation}\label{forward_diffa}
q(\boldsymbol{x}_{t}|\boldsymbol{x}_{t-1}):= \mathcal{N}(\boldsymbol{x}_{t-1}\sqrt{1-b_{t}}, b_{t}\boldsymbol{I}).
\end{equation}
Such a process transforms the data distribution into a standard multivariate normal distribution.
If we denote $a_t=1-b_t$ and $\bar{a}_t=\prod_{s=1}^t a_s$, then we can write:
\begin{equation}\label{conditional_forward_prob}
q(\boldsymbol{x}_{t}|\boldsymbol{x}_{0})= \mathcal{N}(\boldsymbol{x}_{0}\sqrt{\bar{a}_t}, (1-\bar{a}_t)\boldsymbol{I}).
\end{equation}
The reverse process conditioned on the initial sample is also described by a chain of Gaussian distributions:
\begin{equation}\label{conditional_reverse_prob}
q(\boldsymbol{x}_{t-1}|\boldsymbol{x}_{t}, \boldsymbol{x}_{0})= \mathcal{N}(\mu(\boldsymbol{x}_{t}, \boldsymbol{x}_{0}), \boldsymbol{\Sigma}(t)),
\end{equation} where
 \begin{equation*}
\mu(\boldsymbol{x}_{t}, \boldsymbol{x}_{0})=\mu(\boldsymbol{x}_{t}, \boldsymbol{x}_{0}(\boldsymbol{x}_{t}, \boldsymbol{\varepsilon}))=\frac{1}{\sqrt{a_t}}\big(\boldsymbol{x}_{t}- \frac{b_t}{\sqrt{1-\bar{a}_t}}\boldsymbol{\varepsilon} \big),
\end{equation*}
 \begin{equation*}
\boldsymbol{\Sigma}(t)=b_t\boldsymbol{I}.
\end{equation*}
The goal is to approximate this reverse process via
\begin{equation}\label{reverse_diffa}
p(\boldsymbol{x}_{t-1}|\boldsymbol{x}_{t}):= \mathcal{N}(\mu_{\boldsymbol{\theta}}( \boldsymbol{x}_{t},t), \boldsymbol{\Sigma}(t))
\end{equation}
 that converts the standard multivariate normal distribution into the data distribution. To this end, for each step $t$, the KL divergence between $q(\boldsymbol{x}_{t-1}|\boldsymbol{x}_{t}, \boldsymbol{x}_{0})$ and $p(\boldsymbol{x}_{t-1}|\boldsymbol{x}_{t})$ is minimized, which amounts to minimizing
\begin{equation}
\mathbb{E}_{\boldsymbol{x_{0}},\boldsymbol{x_{t}}}||\mu_{\boldsymbol{\theta}}( \boldsymbol{x}_{t},t)-\mu(\boldsymbol{x}_{0}, \boldsymbol{x}_{t})||^2,
\end{equation}  
  or equivalently 
\begin{equation}
\mathbb{E}_{\boldsymbol{x}_{0},\boldsymbol{\varepsilon}}||\boldsymbol{\varepsilon}_{\theta}(\boldsymbol{x_{0}}\sqrt{\bar{a}_t}+\sqrt{(1-\bar{a}_t)}\boldsymbol{\varepsilon},t)-\boldsymbol{\varepsilon}||^2,
\end{equation}  
for $\boldsymbol{\varepsilon} \sim \mathcal{N}(\boldsymbol{0},\boldsymbol{I})$, data samples $\boldsymbol{x_{0}}$, and a neural network $\boldsymbol{\varepsilon}_{\theta}(\boldsymbol{x_{t}},t)$ \citep{DBLP:journals/corr/abs-2006-11239}.
\subsection{SDE Diffusion Models and their CNF Representation}

For an $\boldsymbol{\varepsilon} \sim \mathcal{N}(\boldsymbol{0},\boldsymbol{I})$, the forward diffusion process defined in Equation \ref{forward_diffa} can be written as 
\begin{equation}\label{forward_diff2a}
\boldsymbol{x}_{t}=\boldsymbol{x}_{t-1}\sqrt{1-b_{t}}+ \sqrt{b_{t}} \boldsymbol{\varepsilon}.
\end{equation}
As derived in \citep{song2021scorebased}, the continuous counterpart of this process takes the form
\begin{equation}\label{cont_forward_diffa}
d\boldsymbol{x}(t)=-\frac{1}{2}b(t)\boldsymbol{x}(t)dt+ \sqrt{b(t)} d\boldsymbol{w}.
\end{equation}
In this case Equation \ref{conditional_forward_prob} becomes 
\begin{equation}\label{cont_conditional_forward_probA}
q(\boldsymbol{x}_{t}|\boldsymbol{x}_{0})= \mathcal{N}(\boldsymbol{x}_{0}\mu_t, \sigma_t^2 \boldsymbol{I}),
\end{equation}
where $\mu_t=e^{-\frac{1}{2}\int_0^t b(s)ds}$ and $\sigma_t=\sqrt{(1-e^{-\int_0^t b(s)ds})}$.

As shown through the Fokker-Plack equation, the evolution of the probability density function of the data, as dictated by the FDP, is identical to the evolution dictated by the following ODE transformation:
\begin{equation}\label{smooth_forward_diffa}
d\boldsymbol{x}(t)=-\frac{1}{2}b(t)[\boldsymbol{x}(t)+\nabla_{\boldsymbol{x}(t)}\log{p_t(\boldsymbol{x}(t))}]dt=f_t(\boldsymbol{x}(t))dt.
\end{equation}
Knowing $f_t$ allows us to perform data generation and likelihood estimation, through the framework of continuous normalizing flows. It can be observed that the only unknown in $f_t$, is the score $\nabla_{\boldsymbol{x}(t)}\log{p_t(\boldsymbol{x}(t))}$. This quantity can be modelled using a neural network $\boldsymbol{s}_{\theta}(\boldsymbol{x}(t),t)$ trained either through sliced score matching \cite{DBLP:journals/corr/abs-1905-07088}:
\begin{equation}
min\ \mathbb{E}_{\boldsymbol{x}_t\sim p(\boldsymbol{x}_t)}[\frac{1}{2}||\boldsymbol{s}_{\theta}(\boldsymbol{x}_t,t)||^2+div(\boldsymbol{s}_{\theta}(\boldsymbol{x}_t,t))]
\end{equation}
or the equivalent MSE denoising loss \cite{DBLP:journals/corr/abs-2006-11239, DBLP:journals/corr/abs-2107-00630}:
\begin{equation}
min\ \mathbb{E}_{\boldsymbol{x}_{0},\boldsymbol{\varepsilon}}||\boldsymbol{s}_{\theta}(\boldsymbol{x}_{0}\mu_t+\sigma_t \boldsymbol{\varepsilon},t)-(-\frac{\boldsymbol{\varepsilon}}{\sigma_t})||^2.
\end{equation}  

\begin{algorithm}[H]
   \caption{Standard Approach in Diffusion Models}
   \label{TDUM}
\begin{algorithmic}
   \STATE {\bfseries Input:} data $\{\boldsymbol{x}_1,...,\boldsymbol{x}_n\}$, batch-size p
   \STATE \emph{Train model $\boldsymbol{s}_{\boldsymbol{\theta}}(t)$}
   \REPEAT
   \STATE sample $\tau_1$, ..., $\tau_p$ from $[0,1]$
   \STATE sample $\boldsymbol{x}^0_{\pi(1)}$, ..., $\boldsymbol{x}^0_{\pi(p)}$ from $\{\boldsymbol{x}_1,...,\boldsymbol{x}_n\}$
   \STATE generate $\boldsymbol{x}^{\tau_1}_{\pi(1)}$, ..., $\boldsymbol{x}^{\tau_p}_{\pi(p)}$ using Equation \ref{cont_conditional_forward_probA}
   \STATE minimize $ \Sigma_j ||\boldsymbol{\varepsilon}_{\boldsymbol{\theta}}(\boldsymbol{x}^{\tau_j}_{\pi(j)}(\boldsymbol{\varepsilon_j}),\tau_j)-(\boldsymbol{\varepsilon_j})||^2.$
   \UNTIL convergence

\end{algorithmic}
\end{algorithm}

\section{DDPM and Flow Matching on 2D toy data}\label{AppendixF}
Beyond the investigations delineated in the main paper's experimental section, we conducted additional experiments involving the utilization of 10 blocks that evenly distribute the diffusion process time interval. These experiments were executed within the DDPM framework, which is consistently utilized throughout the paper, but also in this special case in the novel Flow-Matching framework.
\begin{table}[ht]
    \caption{A comparison of the properties of SA-DPM with TPSM$_{10blocks}$, in the DDPM framework. The results are given in negative log-likelihood (lower is better).}
    \label{tableDDPM}
    \centering
    \begin{tabular}{llll}
        \toprule
        Method: & SA-DPM & TPSM$_{10blocks}$ \\
        \midrule
        \makecell{TY}     & 0.76 (1$\times$ \textbf{4H}) & 0.69 (10$\times$ \textbf{1H})  \\
        \makecell{HG}     & 1.11 (1$\times$ \textbf{1.3H}) & 1.05 (10$\times$ \textbf{0.3H}) \\
        \bottomrule
    \end{tabular}
    %\vspace{-10pt}
\end{table}

\begin{figure}[ht]
\center
\includegraphics[width=0.8\textwidth]{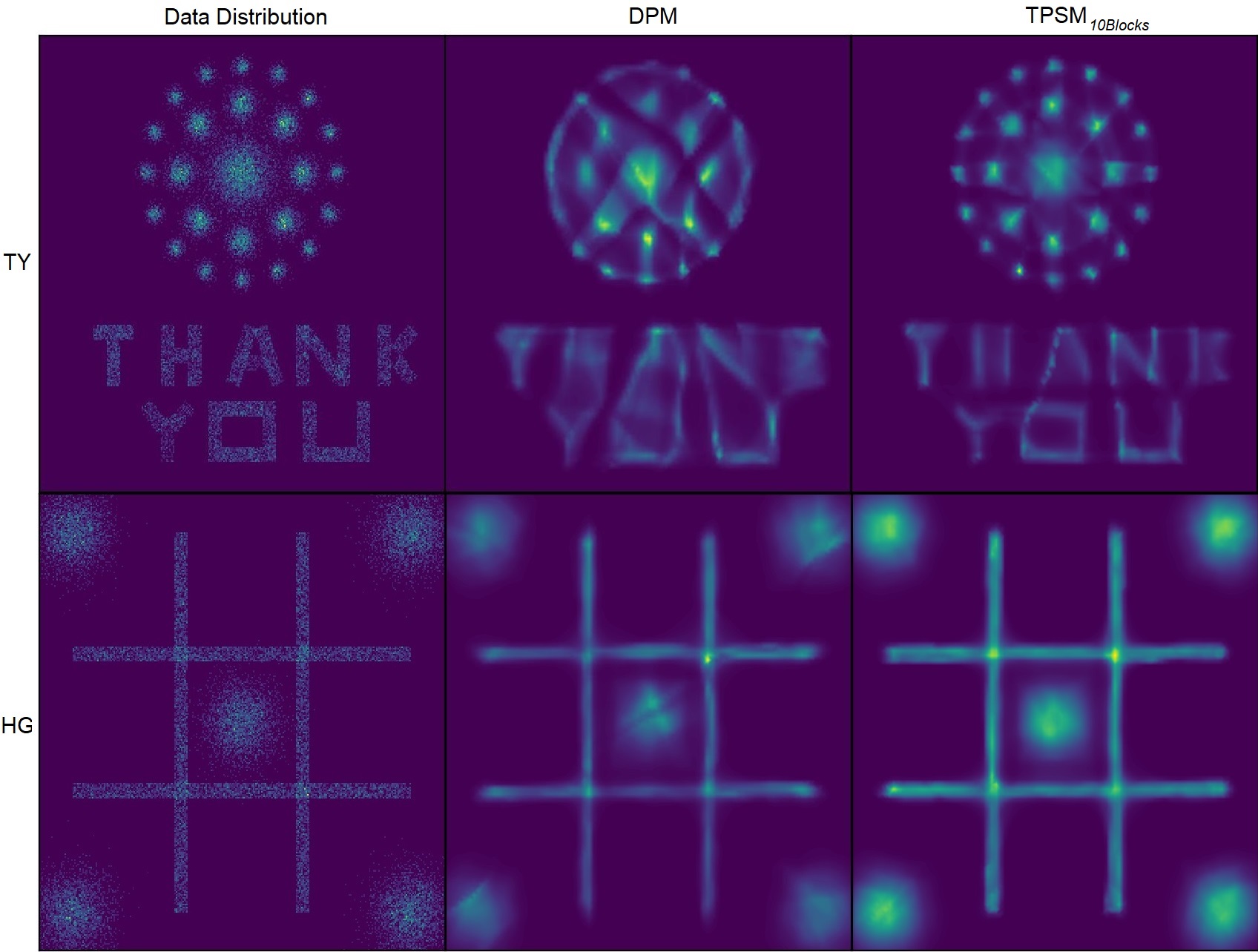}
\caption{SA-DPM and TPSM with 10 blocks comparison on 2D toy data when the DDPM framework is used.}
\label{toy2dimageddpm}
\end{figure}

\begin{figure}[ht]
\center
\includegraphics[width=0.8\textwidth]{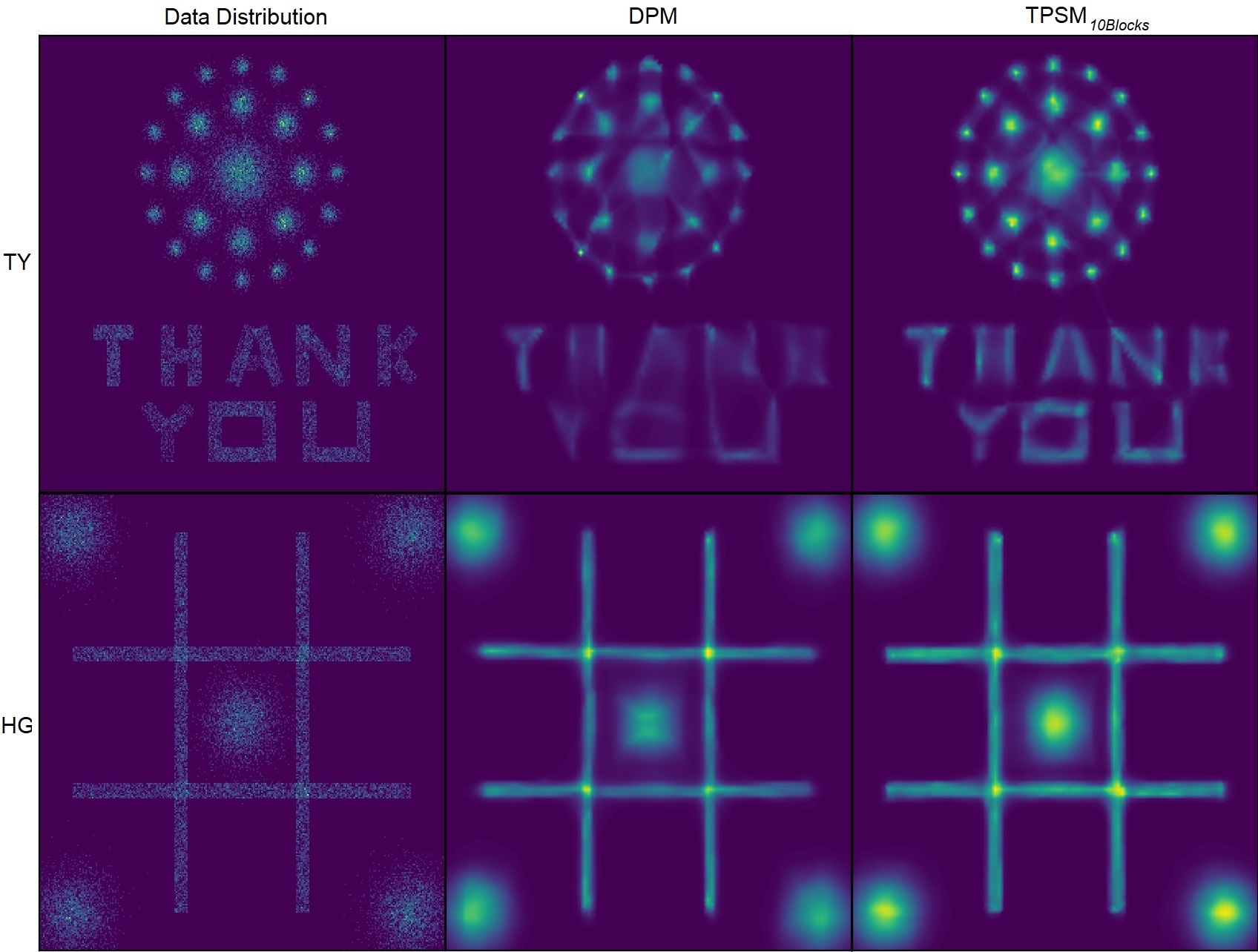}
\caption{SA-DPM and TPSM with 10 blocks comparison on 2D toy data when the OT Flow-Matching framework is used.}
\label{toy2dimagefm}
\end{figure}

Within the DDPM framework, the majority of score (probability) evolution occurs when $t$ approaches 0. Consequently, the outcomes of TPSM$_{10Blocks}$ hold potential for substantial enhancement if more recent settings were employed, such as Flow-Matching, as elucidated in \cite{lipman2023flow}, wherein the evolution of the score is more uniformly distributed over time. This improvement is demonstrated in Table \ref{tableOT} and Figure \ref{toy2dimagefm}. Our observations indicate that the implementation of the Flow-Matching framework bolsters the effectiveness of all approaches. However, the TPSM results exhibit a more pronounced enhancement in the context of the TY data, while in the case of HG the improvements of SA-DPM and TPSM are similar. Thus, even within the confines of more contemporary and efficient frameworks, our parallel score matching strategy continues to deliver the advantageous outcomes postulated in the paper.

\begin{table}[ht]
    \caption{A comparison of the properties of SA-DPM with TPSM$_{10blocks}$, in the OT Flow-Matching framework. The results are given in negative log-likelihood (lower is better).}
    \label{tableOT}
    \centering
    \begin{tabular}{llll}
        \toprule
        Method: & SA-DPM & TPSM$_{10blocks}$ \\
        \midrule
        \makecell{TY}     & 0.75 (1$\times$ \textbf{4H}) & 0.66 (10$\times$ \textbf{1H})  \\
        \makecell{HG}     & 1.10 (1$\times$ \textbf{1.3H}) & 1.04 (10$\times$ \textbf{0.3H}) \\
        \bottomrule
    \end{tabular}
    %\vspace{-10pt}
\end{table}

\newpage
\section{TPSM-B Visual Results}\label{AppendixG}
Owing to the limitations on available space within the primary manuscript, we present here the visual results of TPSM$_B$, which utilizes 4 neural networks. We notice that even if we train these four networks sequentially, the total training time is the same as in SA-DPM, while the resulting performance shows significant improvements. 
\begin{figure}[ht]
\center
\includegraphics[width=0.8\textwidth]{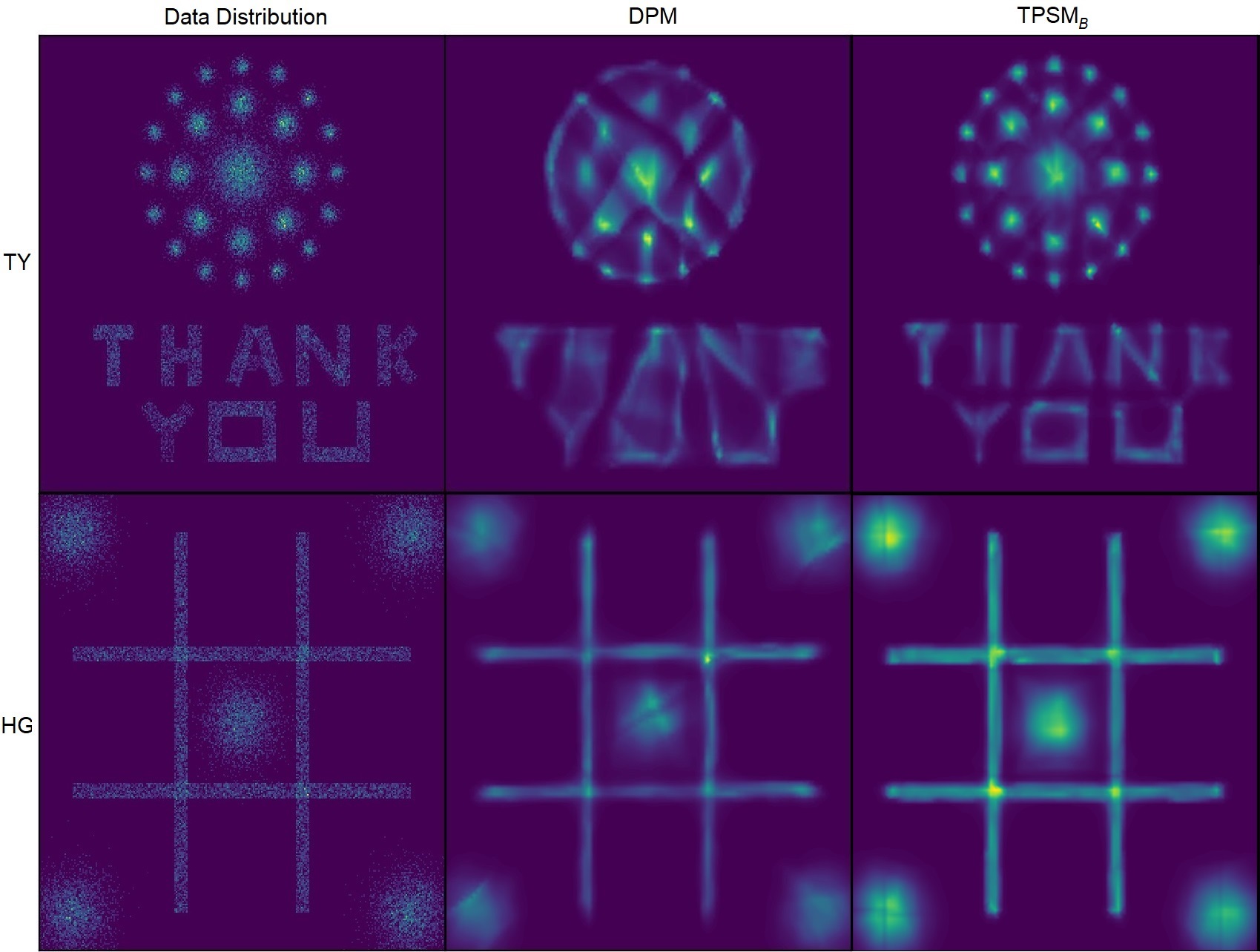}
\caption{TPSM$_B$ visual results compared with SA-DPM.}
\label{2dlkhb}
\end{figure}

\end{document}